\documentclass[%
 reprint,
 amsmath,amssymb,
prb,
]{revtex4-2}

\usepackage{lipsum}
\usepackage{bm}
\usepackage{amssymb}
\usepackage{amstext}
\usepackage{amsmath}
\usepackage{mathrsfs}
\usepackage{graphicx}
\usepackage{amsthm}
\usepackage{floatrow}
\usepackage{url}
\usepackage{listings}
\usepackage{xcolor}
\usepackage{array}

\usepackage{listings}
\usepackage{booktabs}

\definecolor{codegreen}{rgb}{0,0.6,0.4}
\definecolor{codegray}{rgb}{0.5,0.5,0.5}
\definecolor{codepurple}{rgb}{0.58,0,0.82}
\definecolor{backcolour}{rgb}{0.97,0.97,0.97}
\definecolor{keywordcolor}{rgb}{0,0,0.8}

\lstdefinestyle{mystyle}{
    backgroundcolor=\color{backcolour},   
    commentstyle=\color{codegreen},
    keywordstyle=\color{keywordcolor},
    numberstyle=\tiny\color{codegray},
    stringstyle=\color{codepurple},
    basicstyle=\ttfamily\footnotesize,
    breakatwhitespace=false,         
    breaklines=true,                 
    captionpos=b,                    
    keepspaces=true,                 
    numbers=left,                    
    numbersep=5pt,                  
    showspaces=false,                
    showstringspaces=false,
    showtabs=false,                  
    tabsize=2
}
\usepackage{hyperref}

\begin{document}
\preprint{APS/123-QED}
\title{KoopmanLab: machine learning for solving complex physics equations}
\thanks{Correspondence should be addressed to P.S. and Y.T.}%

  \author{Wei Xiong}
\email{xiongw21@mails.tsinghua.edu.cn}
 \altaffiliation[]{Department of Earth System Science, Tsinghua University, Beijing, 100084, China.}
 \author{Muyuan Ma}
\email{mamy22@mails.tsinghua.edu.cn}
 \altaffiliation[]{Department of Earth System Science, Tsinghua University, Beijing, 100084, China.}
 
    \author{Xiaomeng Huang}
   \email{hxm@tsinghua.edu.cn}
 \altaffiliation[]{Department of Earth System Science, Tsinghua University, Beijing, 100084, China.}
 
  \author{Ziyang Zhang}
\email{zhangziyang11@huawei.com}
 \altaffiliation[]{Laboratory of Advanced Computing and Storage, Central Research Institute, 2012 Laboratories, Huawei Technologies Co. Ltd., Beijing, 100084, China.}
\author{Pei Sun}%
 \email{peisun@tsinghua.edu.cn}
 \altaffiliation[]{Department of Psychology \& Tsinghua Brain and Intelligence Lab, Tsinghua University, Beijing, 100084, China.}
 \author{Yang Tian}
\email{tiany20@mails.tsinghua.edu.cn}
 \altaffiliation[]{Department of Psychology \& Tsinghua Laboratory of Brain and Intelligence, Tsinghua University, Beijing, 100084, China.}
 \altaffiliation[Also at]{Laboratory of Advanced Computing and Storage, Central Research Institute, 2012 Laboratories, Huawei Technologies Co. Ltd., Beijing, 100084, China.}



\begin{abstract}
Numerous physics theories are rooted in partial differential equations (PDEs). However, the increasingly intricate physics equations, especially those that lack analytic solutions or closed forms, have impeded the further development of physics. Computationally solving PDEs by classic numerical approaches suffers from the trade-off between accuracy and efficiency and is not applicable to the empirical data generated by unknown latent PDEs. To overcome this challenge, we present KoopmanLab, an efficient module of the Koopman neural operator family, for learning PDEs without analytic solutions or closed forms. Our module consists of multiple variants of the Koopman neural operator (KNO), a kind of mesh-independent neural-network-based PDE solvers developed following dynamic system theory. The compact variants of KNO can accurately solve PDEs with small model sizes while the large variants of KNO are more competitive in predicting highly complicated dynamic systems govern by unknown, high-dimensional, and non-linear PDEs. All variants are validated by mesh-independent and long-term prediction experiments implemented on representative PDEs (e.g., the Navier-Stokes equation and the Bateman–Burgers equation in fluid mechanics) and ERA5 (i.e., one of the largest high-resolution global-scale climate data sets in earth physics). These demonstrations suggest the potential of KoopmanLab to be a fundamental tool in diverse physics studies related to equations or dynamic systems.

\end{abstract}

\maketitle

\section{Introduction}\label{Sec.1}
\subsection{The rising of partial differential equation solvers}
Solving partial differential equations (PDEs) essentially requires characterizing an appropriate solution operator $\mathcal{F}$ that relates $\Phi=\Phi\left(D;\mathbb{R}^{d_{\phi}}\right)$, a Banach space of inputs (i.e., initial values), with $\Gamma=\Gamma\left(D;\mathbb{R}^{d_{\gamma}}\right)$, a Banach space of solutions (i.e., target values),
for a typically time-dependent PDE family defined on a bounded open set $D\subset\mathbb{R}^{d}$
\begin{align}
\partial_{t}\gamma\left(x_{t}\right)&=\left(\mathcal{L}_{\phi}\gamma\right)\left(x_{t}\right)+\kappa\left(x_{t}\right),\;x_{t}\in D\times T,\label{EQ1}\\
\gamma\left(x_{t}\right)&=\gamma_{B},\;x_{t}\in\partial D\times T,\label{EQ2}\\
\gamma\left(x_{0}\right)&=\gamma_{I},\;x_{0}\in D\times \{0\}.\label{EQ3}
\end{align}
In Eq. (\ref{EQ1}-\ref{EQ3}), set $T=\left[0,\infty\right)$ denotes the time domain. Notion $\mathcal{L}_{\phi}$ is a differential operator characterized by $\phi$. Mapping $\kappa\left(\cdot\right)$ is a function that lives in a function space determined by $\mathcal{L}_{\phi}$. Mapping $\gamma\left(\cdot\right)$ is the solution of the PDE family that we attempt to obtain. The boundary and initial conditions are denoted by $\gamma_{B}$ and $\gamma_{I}$, respectively. Mathematically, driving an accurate solution operator $\mathcal{F}:\left(\phi,\gamma_{B},\gamma_{I}\right)\mapsto\gamma$ is the key step to obtain the PDE solution $\gamma\left(\cdot\right)$. However, even in the case where the boundary and initial conditions are constant (i.e., the solution operator $\mathcal{F}:\left(\phi,\gamma_{B},\gamma_{I}\right)\mapsto\gamma$ reduces to $\mathcal{F}:\phi\mapsto\gamma$), driving an analytic expression of solution operator $\mathcal{F}$ can be highly non-trivial \cite{gockenbach2005partial,tanabe2017functional}.

The absence of analytic solutions of various important PDEs in science and engineering naturally calls for the rapid development of computational solvers, which attempt to approximate a parametric counterpart $\mathcal{F}_{\theta}\simeq \mathcal{F}$ parameterized by $\theta$ to derive solution $\gamma\left(\cdot\right)$ \cite{gockenbach2005partial,debnath2005nonlinear,mattheij2005partial}. To date, the joint efforts of physics, mathematics, and computer science have given birth to two mainstream families of PDE solvers \cite{li2020neural}:
\begin{itemize}
    \item[(1)] The first family of solvers are classic numerical ones. Typical instances of these solvers include finite element (FEM) \cite{reddy2019introduction}, finite difference (FDM) \cite{lipnikov2014mimetic}, and finite volume (FVM) \cite{eymard2000finite} methods. In general, these methods discretize space and time domains following specific mesh designs and solve parameterized PDEs on meshes by certain iterative algorithms. Specifically, FEM subdivides the original domain into a set of sub-domains defined by a collection of element equations and recombines these element equations to derive the global solution \cite{reddy2019introduction}. FDM approximates derivatives as finite differences measured on local values \cite{lipnikov2014mimetic}. FVM transforms the original problem into a series of surface flux calculations on local volumes \cite{eymard2000finite}. 
    \item[(2)] The second family of solvers are neural-network-based ones. With a pursuit of accelerating PDE solving and improving the applicability on real data, three kinds of neural-network-based solvers have been proposed:
    \begin{itemize}
        \item[(a)] One kind of solvers discretize domains $D$ and $T$ into $x$ and $y$ meshes and approximate a finite-dimensional and mesh-dependent solution operator $\mathcal{F}_{\theta}$ by a parameterized neural network between finite Euclidean spaces, i.e., $\mathcal{F}_{\theta}:\mathbb{R}^{x}\times\mathbb{R}^{y}\times\Theta\rightarrow\mathbb{R}^{x}\times\mathbb{R}^{y}$ (e.g., see Refs. \cite{guo2016convolutional,zhu2018bayesian,bhatnagar2019prediction}). Given an arbitrary input $\gamma\left(x_{t}\right)$, the trained neural network can function as a solution operator to predict $\gamma\left(x_{t+\tau}\right)=\mathcal{F}_{\theta}\left(\gamma\left(x_{t}\right)\right)$ for a certain time difference $\tau$.
        \item[(b)] Another kind of solvers directly parameterize equation solution $\gamma\left(\cdot\right)$ as a neural network, i.e., $\mathcal{F}_{\theta}:D\times T\times\Theta\rightarrow\mathbb{R}$ (e.g., see Refs. \cite{yu2018deep,raissi2019physics,bar2019unsupervised,pan2020physics}). These solvers are mesh-independent and accurate in learning a given PDE because they can directly transform arbitrary domain and parameter setting to target equation solution $\gamma\left(\cdot\right)$.
        \item[(c)] The last kind of solvers, including neural operators, attempt to parameterize a mesh-dependent and infinite-dimensional solution operator with neural networks, i.e, $\mathcal{F}_{\theta}:\Phi\times\Theta\rightarrow\Gamma$ (e.g., see Refs. \cite{lu2019deeponet,bhattacharya2020model,nelsen2021random,li2020neural,li2020fourier,kovachki2021universal,li2022fourier}). These mesh-independent solvers can be flexibly implemented on different discretization schemes and only need to be trained once for a given PDE family. The equation solution $\gamma\left(\cdot\right)$ of different instances of the PDE family can be generated by a computationally reusable forward pass of the network \cite{li2020neural,li2020fourier}, which can be further accelerated by fast Fourier transform \cite{li2020fourier}. Representative demonstrations of this kind of solver are Fourier neural operator \cite{li2020fourier} and its variants (e.g., adaptive Fourier neural operator \cite{guibas2021efficient} and FourCastNet \cite{pathak2022fourcastnet,kurth2022fourcastnet}). These frameworks not only solve PDEs with known expressions but also be able to predict complex dynamic systems governed by unknown PDEs on real data sets (e.g., climate system \cite{pathak2022fourcastnet,kurth2022fourcastnet}).
    \end{itemize}
\end{itemize}

\subsection{The limitation of previous partial differential equation solvers}

Although substantial progress has been accomplished by existing PDE solvers from various perspectives, there remain critical challenges in this booming direction.

In practice, the mesh-dependent property of classic numerical solvers has implied an inevitable trade-off between computation accuracy and efficiency, i.e., fine-grained meshes ensure accuracy yet coarse-grained meshes are favorable for efficiency \cite{li2020fourier,tadmor2012review}. However, in many cases, the applications of PDE solving (e.g., numerical weather forecasting \cite{stensrud2009parameterization,bauer2015quiet}) require timely and accurate computation. To ensure accuracy and speed, every single time of computation in the downstream applications supported by classic numerical solvers frequently costs large amounts of computing resources. In cases with limited computing power, a significant time delay may occur. Moreover, all numerical solvers require the explicit definitions of target PDEs as \emph{a priori} knowledge and are less applicable to predict real data generated by unknown PDEs \cite{li2020fourier}.

As for neural-network-based solvers, challenges still arise from multiple perspectives, even though these solvers have outperformed the classic numerical ones in prediction efficiency significantly. Type (a) solvers, as we have suggested, are mesh-dependent and lack generalization capacities across different mesh designs \cite{li2020neural}. Type (b) solvers are limited to learning a concrete instance of the PDE rather than the entire family and, consequently, require restarted training given a different instance and can not handle the data with unknown PDEs \cite{li2020neural}. Although type (c) solvers can learn the entire PDE family in a mesh-independent manner \cite{li2020fourier,li2020neural}, they may face challenges in characterizing the long-term behaviour of equation solution $\gamma\left(\cdot\right)$. To understand these challenges, let us consider the iterative update strategy of neural operators for any $x_{t}\in D\times \{t\}$ \cite{li2020neural}
\begin{align}
&\widehat{\gamma}\left(x_{t+\varepsilon}\right)\notag\\=&\sigma\left(W\widehat{\gamma}\left(x_{t}\right)+\int_{D\times \{t\}}\kappa_{\theta}\left(x_{t},y_{t},\phi\left(x_{t}\right),\phi\left(y_{t}\right)\right)\widehat{\gamma}\left(y_{t}\right)\mathsf{d}y_{t}\right),\label{EQ4}
\end{align}
in which $\varepsilon\in\left(0,\infty\right)$ denotes time difference, notion $\sigma:\mathbb{R}\rightarrow\mathbb{R}$ is an arbitrary element-wise non-linear activation function, notion $W:\mathbb{R}^{d_{\widehat{\gamma}}}\rightarrow\mathbb{R}^{d_{\widehat{\gamma}}}$ stands for a linear layer, function $\kappa_{\theta}:\mathbb{R}^{2\left(d+d_{\phi}\right)}\rightarrow\mathbb{R}^{d_{\widehat{\gamma}}}$ is a neural network parameterized by $\theta$, and mapping $\widehat{\gamma}:D\times T\rightarrow\mathbb{R}^{d_{\widehat{\gamma}}}$ denotes the parameterized counterpart of equation solution $\gamma$ generated by the neural network (e.g., by embedding) \cite{li2020neural}. In Eq. (\ref{EQ4}), the integral term associated with $\kappa_{\theta}$ defines an kernel integral operator to parameterize the Green function $\mathcal{J}_{\phi}:\left(D\times T\right)\times \left(D\times T\right)\rightarrow \mathbb{R}$
\begin{align}
&\widehat{\gamma}\left(x_{t+\varepsilon}\right)=\int_{D\times \{t\}}\mathcal{J}_{\phi}\left(x_{t},y_{t}\right)\eta\left(y_{t}\right)\mathsf{d}y_{t},\;\forall \;x_{t}\in D\times \{t\},\label{EQ5}
\end{align}
where the Green function is determined by $\phi$ as well. One can see a similar form of Eq. (\ref{EQ5}) in Ref. \cite{li2020neural}. Computationally, the iteration of Eq. (\ref{EQ4}) can be significantly accelerated by Fourier transform, which leads to the well-known Fourier neural operator \cite{li2020fourier}.

From a dynamic system perspective, Eq. (\ref{EQ4}) is similar to the iterative dynamics of an infinite-dimensional non-linear dynamic system of equation solution $\gamma_{t}=\gamma\left(D\times\{t\}\right)$, where each snapshot $\gamma\left(D\times\{t\}\right)$ is generated after function $\gamma$ acts on all elements in set $D\times\{t\}$. Mathematically, the dynamics is defined as
\begin{align}
\gamma_{t+\varepsilon}=\gamma_{t}+\int_{t}^{t+\varepsilon}\zeta\left(\gamma_{\tau},\tau\right)\mathsf{d}\tau,\;\forall t\in T,\label{EQ6}
\end{align}
or equivalently
\begin{align}
\partial_{t}\gamma_{t}=\zeta\left(\gamma_{t},t\right),\;\forall\gamma_{t}\in\mathbb{R}^{d_{\gamma}}\times T,\label{EQ7}
\end{align}
in which $\zeta:\mathbb{R}^{d_{\gamma}}\times T\rightarrow\mathbb{R}^{d_{\gamma}}$ denotes the associated infinite-dimensional evolution mapping.

The challenge faced by type (c) solvers lies in that evolution mapping $\zeta\left(\cdot,\cdot\right)$ maybe even more intricate than equation solution $\gamma\left(\cdot\right)$ itself. Let us consider the cocycle property of the flow mapping $\theta$ associated with $\zeta\left(\cdot,\cdot\right)$ according to modern dynamic system theory \cite{brunton2022modern}
\begin{align}
\theta_{t}^{t+\varepsilon}=\theta_{t+\tau}^{t+\varepsilon}\circ\theta_{t}^{t+\tau},\;\forall t\leq t+\tau\leq t+\varepsilon\in T.\label{EQ8}
\end{align}
Operator $\circ$ denotes the composition of mappings. In general, Eq. (\ref{EQ8}) determines how equation solution $\gamma\left(\cdot\right)$ evolves across adjoining time intervals. In a special case where $\zeta\left(\cdot,\cdot\right)$ is time-independent, i.e., $\partial_{t}\zeta\left(\cdot,t\right)\equiv 0$, Eq. (\ref{EQ8}) reduces to the autonomous case 
\begin{align}
\theta^{t+\varepsilon}=\theta^{\varepsilon}\circ\theta^{t},\;\forall t,\varepsilon\in T.\label{EQ9}
\end{align}
Otherwise, Eq. (\ref{EQ8}) generally corresponds to the non-autonomous case where the underlying mechanisms governing the evolution of $\gamma\left(\cdot\right)$ vary across time. Consequently, a large $\varepsilon$ may correspond to a highly non-trivial evolution process of $\gamma\left(\cdot\right)$, making $\widehat{\gamma}\left(x_{t+\varepsilon}\right)$ less predictable during iterative updating and reducing the precision of Eq. (\ref{EQ4}) significantly. This phenomenon inevitably impedes the accurate prediction of the long-term dynamics (i.e., $\varepsilon\rightarrow\infty$) of diverse non-linear PDE families (e.g., see those in epidemic prevention \cite{cao2020spectral}, economic modelling \cite{aminian2006forecasting}, and weather forecast \cite{pathak2022fourcastnet,kurth2022fourcastnet}). To overcome this obstacle, existing models are forced to improve accuracy at the cost of efficiency. 

\subsection{Our contributions to partial differential equation solvers}

In this paper, we build on Koopman neural operator (KNO), one of our latest works \cite{anonymous2023koopman}, to develop an efficient module of PDE solving and overcome the limitation in characterizing the long-term behaviours of complicated PDE families. As a study on computational physics programs, our research has the following contributions compared with our previous work \cite{anonymous2023koopman}.

First, we generalize the original KNO to four kinds of variants. Beyond the original KNO, these differentiated variants offer more possibilities for data-specific and task-oriented solver designs. Specifically, the compact variants of KNO realized by multi-layer perceptrons and convolutional neural networks can accurately solve PDE with small model sizes. The large variants of KNO implemented on visual transformers can predict highly intricate dynamic systems governed by unknown, high-dimensional, and non-linear PDEs (e.g., climate system).  

Second, we propose KoopmanLab, a PyTorch module of Koopman neural operator family, as a self-contained and user-friendly platform for PDE solving. All necessary tools, such as those for data loading, model construction, parameter manipulation, output visualization, and performance quantification, are offered in a user-friendly manner to support customized applications.

Third, we offer comprehensive validation of the proposed module on representative data sets, including those generated by important PDEs in fluid mechanics (e.g., the Navier-Stokes equation and the Bateman–Burgers equation) or obtained by global meteorological recording research (e.g., atmospheric, land, and oceanic climate fields in ERA5 data set) \cite{hersbach2020era5}. By measuring accuracy, quantifying efficiency, and comparing all KNO variants with other state-of-the-art alternatives (e.g., Fourier neural operator \cite{li2020fourier} and FourCastNet \cite{pathak2022fourcastnet,kurth2022fourcastnet}), we suggest the potential of our module to serve as an ideal choice of PDE solving and dynamic system prediction.

\section{The initial version of Koopman neural operator}\label{Sec2}
Although the original Koopman neural operator has been proposed in our earlier work \cite{anonymous2023koopman}, here we elaborate on its mechanisms for completeness. We further present more mathematical details that are not covered in Ref. \cite{anonymous2023koopman} to analyze the convergence of the original Koopman neural operator.

\subsection{The original Koopman neural operator: Objective}
Koopman neural operator (KNO) is proposed to deal with the non-linear, and potentially non-autonomous, dynamic system in Eqs. (\ref{EQ6}-\ref{EQ7}). The idea underlying KNO arises from the pursuit to transform the non-linear system in Eqs. (\ref{EQ6}-\ref{EQ7}) to a sufficiently simple linear one
\begin{align}
\partial_{t}\mathbf{g}\left(\gamma_{t}\right)=\mathcal{A}\mathbf{g}\left(\gamma_{t}\right),\;\forall t\in T,\label{EQ10}
\end{align}
where $\mathbf{g}\left(\cdot\right)$ is an appropriate transform and $\mathcal{A}$ is a linear operator. In modern dynamic system theory \cite{brunton2022modern}, this pursuit may be achieved if we can develop an approach to characterize the Koopman operator $\mathcal{K}$, an infinite-dimensional linear operator governing all possible observations of the dynamic system of equation solution $\gamma\left(\cdot\right)$, to act on the flow mapping $\theta$ and linearizing the dynamics of $\gamma\left(\cdot\right)$ in an appropriate observation space. This idea has been extensively applied in plasma physics \cite{taylor2018dynamic}, fluid dynamics \cite{rowley2009spectral}, robot kinetics \cite{abraham2019active}, and neuroscience \cite{brunton2016extracting}.

Mathematically, we need to find a set of observation functions (or named as measurement functions) \cite{brunton2022modern}
\begin{align}
    \mathcal{G}\left(\mathbb{R}^{d_{\gamma}}\times T\right)=\{\mathbf{g}\vert \mathbf{g}:\mathbb{R}^{d_{\gamma}}\times T\rightarrow\mathbb{C}^{d_{\gamma}}\} \label{EQ11}
\end{align}
such that a family of Koopman operators can be identified for the autonomous (i.e., $\mathcal{K}^{\varepsilon}:\mathcal{G}\left(\mathbb{R}^{d_{\gamma}}\times T\right)\rightarrow\mathcal{G}\left(\mathbb{R}^{d_{\gamma}}\times T\right)$) or the non-autonomous (i.e., $\mathcal{K}^{t+\varepsilon}_{t}:\mathcal{G}\left(\mathbb{R}^{d_{\gamma}}\times T\right)\rightarrow\mathcal{G}\left(\mathbb{R}^{d_{\gamma}}\times T\right)$) case. These Koopman operators can function on the observations of $\gamma\left(\cdot\right)$ to update them
\begin{align}
\mathcal{K}^{\varepsilon}\mathbf{g}\left(\gamma_{t}\right)&=\mathbf{g}\left(\theta^{\varepsilon}\left(\gamma_{t}\right)\right)=\mathbf{g}\left(\gamma_{t+\varepsilon}\right),\;\forall t\times T,\label{EQ12}\\
\mathcal{K}^{t+\varepsilon}_{t}\mathbf{g}\left(\gamma_{t}\right)&=\mathbf{g}\left(\theta^{t+\varepsilon}_{t}\left(\gamma_{t}\right)\right)=\mathbf{g}\left(\gamma_{t+\varepsilon}\right),\;\forall t\leq t+\varepsilon\in T,\label{EQ13}
\end{align}
where Eqs. (\ref{EQ12}-\ref{EQ13}) correspond to the autonomous and non-autonomous cases, respectively. The updating is implemented in a linear manner, which can be illustrated by taking the non-autonomous case as an example
\begin{align}
\partial_{t}\mathbf{g}\left(\gamma_{t}\right)=\lim_{\varepsilon\rightarrow0}\frac{\mathcal{K}^{t+\varepsilon}_{t}\mathbf{g}\left(\gamma_{t}\right)-\mathbf{g}\left(\gamma_{t}\right)}{\varepsilon}.\label{EQ14}
\end{align}
Apart from the linear system of $\mathbf{g}\left(\gamma_{t}\right)$ in Eq. (\ref{EQ14}), one may also consider the Lie operator (i.e., the Lie derivative of $\mathbf{g}\left(\cdot\right)$ along the vector field $\gamma\left(\cdot\right)$), which is generator operator of such a Koopman operator \cite{koopman1931hamiltonian,abraham2012manifolds,chicone1999evolution}
\begin{align}
\mathcal{L}_{t}\mathbf{g}=\lim_{t+\varepsilon\rightarrow t}\frac{\mathcal{K}^{t+\varepsilon}_{t}\mathbf{g}\left(\gamma_{t}\right)-\mathbf{g}\left(\gamma_{t}\right)}{t+\varepsilon-t}.\label{EQ15}
\end{align}
Eq. (\ref{EQ15}) defines a linear system of $\mathbf{g}\left(\gamma_{t}\right)$ as well
\begin{align}
\partial_{t}\mathbf{g}\left(\gamma_{t}\right)=\lim_{t+\varepsilon\rightarrow t}\frac{\mathcal{K}^{t+\varepsilon}_{t}\mathbf{g}\left(\gamma_{t}\right)-\mathbf{g}\left(\gamma_{t}\right)}{\varepsilon}=\mathcal{L}_{t}\mathbf{g}\left(\gamma_{t}\right),\label{EQ16}
\end{align}
which can also be considered in the application.

To understand the linearization of $\mathbf{g}\left(\gamma_{t}\right)$ by the Koopman operator from the perspective of PDE solving, let us consider the Lax pair $\left(\mathcal{M},\mathcal{N}\right)$ of an integrable version of Eqs. (\ref{EQ1}-\ref{EQ3}) \cite{lax1968integrals}
\begin{align}
\mathcal{M}&=\mathsf{D}_{x}^{n}+\alpha\gamma\left(x_{t}\right)I,\;\alpha\in\mathbb{C},\label{EQ17}\\
\mathcal{M}\psi\left(x_{t}\right)&=\lambda\psi\left(x_{t}\right),\;\lambda\in\mathbb{C},\label{EQ18}\\
\partial_{t}\psi\left(x_{t}\right)&=\mathcal{N}\psi\left(x_{t}\right),\label{EQ19}
\end{align}
where $\mathsf{D}_{x}^{n}$ denotes the  $n$-th total derivative operator and $I$ is an identity operator. Eq. (\ref{EQ18}) denotes an eigenvalue problem at moment $t$. A relation between linear operators $\mathcal{M}$ and $\mathcal{N}$ can be identified if we calculate the time derivative of Eq. (\ref{EQ18})
\begin{align}
\left(\partial_{t}\mathcal{M}+\mathcal{M}\mathcal{N}-\mathcal{N}\mathcal{M}\right)\psi\left(x_{t}\right)=\partial_{t}\lambda\psi\left(x_{t}\right),\label{EQ20}
\end{align}
which directly leads to
\begin{align}
\partial_{t}\mathcal{M}+\left[\mathcal{M},\mathcal{N}\right]=0,\label{EQ21}
\end{align}
where $\left[\mathcal{M},\mathcal{N}\right]=\mathcal{M}\mathcal{N}-\mathcal{N}\mathcal{M}$ denotes the commutator of operators. Combining Eqs. (\ref{EQ17}-\ref{EQ21}) with Eq. (\ref{EQ16}), we can readily see the close relation between $\mathcal{N}$ and $\mathcal{K}^{t+\varepsilon}_{t}$
\begin{align}
\psi\left(D\times\{t\}\right)=\mathbf{g}\left(\gamma_{t}\right)\Rightarrow\mathcal{N}=\lim_{t+\varepsilon\rightarrow t}\frac{\mathcal{K}^{t+\varepsilon}_{t}\mathbf{g}\left(\gamma_{t}\right)-\mathbf{g}\left(\gamma_{t}\right)}{\varepsilon},\label{EQ22}
\end{align}
which holds in the autonomous case as well. In sum, the linearization of $\mathbf{g}\left(\gamma_{t}\right)$ is intrinsically related to the Lax pair and the inverse scattering transform of integrable PDEs \cite{lax1968integrals}. Note that similar ideas have been comprehensively explored in mathematics and physics \cite{parker2020koopman,nakao2020spectral,gin2021deep,page2018koopman}.

Once we find a way to derive the Koopman operator, we can reformulate Eq. (\ref{EQ4}) as
\begin{align}
&\widehat{\gamma}_{t+\varepsilon}=\mathbf{g}^{-1}\left[\mathcal{K}^{t+\varepsilon}_{t}\mathbf{g}\left(\widehat{\gamma}_{t}\right)\right],\;\forall t\in T,\label{EQ23}
\end{align}
where $\widehat{\gamma}_{t}=\widehat{\gamma}\left(D\times\{t\}\right)$. Certainly, an infinite-dimensional linear operator is not operable in practice. To enable neural networks to learn a potential Koopman operator, we need to consider $\widehat{\mathcal{K}}\in\mathbb{R}^{r\times r}$, a finite matrix, as a counterpart of $\mathcal{K}$ that acts on $\mathbb{K}=\operatorname{span}\left(\widehat{\mathcal{G}}\right)$, a finite invariant sub-space spanned by $\widehat{\mathcal{G}}=\{\mathbf{g}_{1},\ldots,\mathbf{g}_{r}\}\subset \mathcal{G}\left(\mathbb{R}^{d_{\gamma}}\times T\right)$
\begin{align}
\widehat{\mathcal{K}}\mathbf{g}_{i}=\langle\left[\nu_{1},\ldots,\nu_{r}\right],\left[\mathbf{g}_{1},\ldots,\mathbf{g}_{r}\right]\rangle,\;\forall\mathbf{g}_{i}\in\widehat{\mathcal{G}},\label{EQ24}
\end{align}
where $\left[\nu_{1},\ldots,\nu_{r}\right]\in\mathbb{R}^{r}$ and $\langle\cdot,\cdot\rangle$ denotes the inner product. Mathematically, any finite set of eigenfunctions of the Koopman operator $\mathcal{K}$ can span a finite invariant sub-space.

\subsection{The original Koopman neural operator: Mathematics}
There exist numerous previous works that pursue to characterize the Koopman operator by machine-learning-based approaches. Some approaches are highly practical but limited to the autonomous case (e.g., the case in Eq. (\ref{EQ12})) \cite{takeishi2017learning,azencot2020forecasting,otto2019linearly,alford2022deep}. Other approaches are more general in application but critically depend on \emph{a priori} knowledge about the eigenvalue spectrum (e.g, the numbers of real and complex eigenvalues) of Koopman operator to deal with the continuous spectrum problem \cite{lusch2018deep}.

In practice, a balance should be reached between mathematical completeness and computational practicality. An ideal Koopman-operator-based PDE solver should fit in with both autonomous and non-autonomous cases and limit the dependence of \emph{a priori} knowledge as much as possible (even though these restraints inevitably reduce mathematical completeness). To explore such a balance, we introduce the Koopman neural operator (KNO), a flexible approach, in our previous work \cite{anonymous2023koopman}.

The formalization of KNO begins with the Krylov sequence \cite{saad2011numerical} of the observable defined by a unit time step $\varepsilon\in\left[0,\infty\right]$, which is used in the Krylov subspace approach for computing the eigenvalues of large matrices \cite{saad2011numerical}. One can see its application in Koopman-operator-related algorithms such as the Hankel-DMD \cite{arbabi2017ergodic}, sHankel-DMD \cite{vcrnjaric2020koopman}, and HAVOK \cite{brunton2017chaos}. Specifically, the Krylov sequence is given as 
\begin{align}
\mathcal{R}_{n}&=\left[\mathbf{g}\left(\gamma_{0}\right),\mathbf{g}\left(\gamma_{\varepsilon}\right),\mathbf{g}\left(\gamma_{2\varepsilon}\right),\ldots,\mathbf{g}\left(\gamma_{n\varepsilon}\right)\right],\label{EQ25}
\end{align}
which is generated by $\mathcal{K}$ and $\mathbf{g}\left(\gamma_{0}\right)$
\begin{align}
&\mathcal{R}_{n}\notag\\=&\left[\mathbf{g}\left(\gamma_{0}\right),\mathcal{K}^{\varepsilon}_{0}\mathbf{g}\left(\gamma_{0}\right),\mathcal{K}^{2\varepsilon}_{\varepsilon}\mathcal{K}^{\varepsilon}_{0}\mathbf{g}\left(\gamma_{0}\right),\ldots,\mathcal{K}^{n\varepsilon}_{\left(n-1\right)\varepsilon}\cdots\mathcal{K}^{\varepsilon}_{0}\mathbf{g}\left(\gamma_{0}\right)\right].\label{EQ26}
\end{align}
Computationally, the Krylov sequence can be sampled by a Hankel matrix of observations
\begin{align}
\mathcal{H}_{m\times n}&=\begin{bmatrix} 
	\mathbf{g}\left(\gamma_{0}\right) & \mathbf{g}\left(\gamma_{\varepsilon}\right) & \cdots & \mathbf{g}\left(\gamma_{n\varepsilon}\right) \\
	\vdots & \vdots & \vdots & \vdots\\
	\mathbf{g}\left(\gamma_{\left(m-1\right)\varepsilon}\right) & \mathbf{g}\left(\gamma_{m\varepsilon}\right) & \cdots & \mathbf{g}\left(\gamma_{\left(m+n-1\right)\varepsilon}\right) \\
	\end{bmatrix},\label{EQ27}
\end{align}
where $m\in\mathbb{N}^{+}$ denotes the dimension of delay-embedding. In Eq. (\ref{EQ27}), each column is a sampled result that approximates a function in the Krylov subspace. 

If the Koopman operator has a discrete spectrum (e.g., has eigenvalues), there exists an invariant subspace $\mathbb{K}$ of the Koopman operator, which can be spanned by the Krylov subspace
\begin{align}
\mathbb{K}=\operatorname{span}\left(\mathcal{R}_{n}\right)\simeq\operatorname{span}\left(\mathcal{H}_{\left(m,n\right)}\right)\label{EQ28}
\end{align}
as long as $n\geq \operatorname{dim}\left(\mathbb{K}\right)-1$ (here $\operatorname{dim}\left(\cdot\right)$ denotes the dimensionality). This property suggests the possibility of approximating the actual Koopman operator to $\mathbb{K}$ by $\widehat{\mathcal{K}}_{t}^{t+\varepsilon}:\mathcal{G}\left(\mathbb{R}^{d_{\gamma}}\times T\right)\rightarrow\mathbb{K}$, a finite Koopman operator restricted to $\mathbb{K}$ for any $t\in T$. Mathematically, matrix $\widehat{\mathcal{K}}$ is required to satisfy the Galerkin projection relation 
\begin{align}
\langle\widehat{\mathcal{K}}_{t}^{t+\varepsilon}\mathbf{h}\left(\gamma_{t}\right),\mathbf{g}\left(\gamma_{i\varepsilon}\right)\rangle=\langle\mathcal{K}_{t}^{t+\varepsilon}\mathbf{h}\left(\gamma_{t}\right),\mathbf{g}\left(\gamma_{i\varepsilon}\right)\rangle,\;\forall i=0,\ldots,m,\label{EQ29}
\end{align}
where $\mathbf{h}\left(\cdot\right)\in\mathcal{G}\left(\mathbb{R}^{d_{\gamma}}\times T\right)$ is an arbitrary function \cite{korda2018convergence,li2022reduced}. If the target Koopman operator is bounded and $\mathcal{H}_{\left(m,n\right)}$ spans its invariant subspace, the approximation can be realized by 
\begin{align}
&\lim_{m\rightarrow\infty}\int_{\mathcal{G}\left(\mathbb{R}^{d_{\gamma}}\times T\right)}\Vert \widehat{\mathcal{K}}_{t}^{t+\varepsilon}\mathbf{h}\left(\gamma_{t}\right)-\mathcal{K}_{t}^{t+\varepsilon}\mathbf{h}\left(\gamma_{t}\right)\Vert_{F}\mathsf{d}\mu=0,\notag\\&\forall \mathbf{h}\left(\cdot\right)\in\mathcal{G}\left(\mathbb{R}^{d_{\gamma}}\times T\right), \label{EQ30}
\end{align}
where $\mu$ is a measure on $\mathcal{G}\left(\mathbb{R}^{d_{\gamma}}\times T\right)$ and $\Vert\cdot\Vert_{F}$ denotes the Frobenius norm. Once a restricted Koopman operator is derived, we can obtain the following iterative dynamics 
\begin{align}
    \mathcal{H}_{m\times n}\left(k+1\right)=\widehat{\mathcal{K}}_{k\varepsilon}^{\left(k+1\right)\varepsilon}\mathcal{H}_{m\times n}\left(k\right),\;\forall k=1,\ldots,n, \label{EQ31}
\end{align}
in which $\mathcal{H}_{m\times n}\left(k\right)$ is the $k$-th column of $\mathcal{H}_{m\times n}$. 

As for the case where the Koopman operator has a continuous spectrum (e.g., has no eigenvalue), there is no finite invariant subspace to support computational approximation. Such an ill-posed situation remains for future exploration.

The restricted Koopman operator $\widehat{\mathcal{K}}$ can be learned efficiently if it corresponds to autonomous system, i.e., $\widehat{\mathcal{K}}_{t}^{t+\varepsilon}=\widehat{\mathcal{K}}^{\varepsilon}$. However, an online optimization will be necessary if it corresponds to non-autonomous system, i.e., $\widehat{\mathcal{K}}_{t}^{t+\varepsilon}$ is time-varying. Limited by computing resources or data size, expensive online optimization may not always be available during PDE solving. Therefore, we propose a compromised approach to realize off-line training under the ergodic assumption \cite{arbabi2017ergodic,cornfeld2012ergodic} of the dynamic system of $\gamma_{t}$, i.e., $\gamma_{t}$ ultimately visits every possible system states as $t\rightarrow \infty$. Under this assumption, the proportion of retention time of $\gamma_{t}$ at a certain system state is equivalent to the probability of this state in the space, making the time-averaging equivalent to the actual expectation at the limit of infinite time. Based on this property, we can define an expectation of the restricted Koopman operator associated with $\varepsilon$
\begin{align}
\overline{\mathcal{K}}_{\varepsilon}&=\lim_{t\rightarrow\infty}\frac{1}{t}\int_{\left[0,t\right)}\mathbf{g}\left(\gamma_{\tau}\right)^{-1}\mathbf{g}\left(\gamma_{\tau+\varepsilon}\right)\mathsf{d}\tau,\label{EQ32}\\&\simeq\operatorname*{argmin}_{P\in\mathbb{R}}\sum_{k=1}^{n}\Vert\mathcal{H}_{m\times n}\left(k+1\right)-P\mathcal{H}_{m\times n}\left(k\right)\Vert_{F}.\label{EQ33}
\end{align}
For a fixed time difference $\varepsilon$, the expected Koopman operator $\overline{\mathcal{K}}_{\varepsilon}:\mathcal{G}\left(\mathbb{R}^{d_{\gamma}}\times T\right)\rightarrow\mathbb{K}$ is a time-average of $\widehat{\mathcal{K}}_{t}^{t+\varepsilon}$ that can be learned during offline optimization.

\subsection{The original Koopman neural operator: Convergence}

Given an ideal setting of $m\rightarrow \infty$, we can ensure the convergence of the eigenvalues and eigenfunctions of $\widehat{\mathcal{K}}$ to those of $\mathcal{K}$ under the assumption of ergodic property. Similar conclusions can be seen in Ref. \cite{korda2018convergence}. To understand this convergence, we need to indicate several important properties. First, as we have mentioned, there exists an equivalence relation between time-averaging and the real expectation as the time approaches to infinity (i.e., the Birkhoff ergodic theorem \cite{arbabi2017ergodic,cornfeld2012ergodic})
\begin{align}
\lim_{m\rightarrow\infty}\frac{1}{m}\sum_{i=0}^{m-1}\mathbf{g}\left(\gamma_{i}\right)=\int_{\mathbb{K}}\mathbf{g}\mathsf{d}\mu.\label{EQ34}
\end{align}
Second, Eq. (\ref{EQ34}) directly implies that
\begin{align}
&\lim_{m\rightarrow\infty}\frac{1}{m}\langle\mathcal{H}_{m\times n}\left(i\right),\mathcal{H}_{m\times n}\left(j\right)\rangle\notag\\=&\int_{\mathbb{K}}\mathcal{H}_{m\times n}\left(i\right)\left[\mathcal{H}_{m\times n}\left(j\right)\right]^{*}\mathsf{d}\mu.\label{EQ35}\\
=&\langle\mathcal{H}_{m\times n}\left(i\right),\mathcal{H}_{m\times n}\left(j\right)\rangle_{\mathbb{K}},\label{EQ36}
\end{align}
where $*$ denotes the complex conjugate and $\langle\cdot,\cdot\rangle_{\mathbb{K}}$ stands for the inner product of functions in $\mathbb{K}$. Given the learned Koopman operator $\overline{\mathcal{K}}_{\varepsilon}$, Eq. (\ref{EQ36}) coincides with the definition of the actual Gramian matrix $\mathcal{V}$ associated with the inner product space $\mathbb{K}$
\begin{align}
\mathcal{V}_{i,j}&=\langle\overline{\mathcal{K}}_{\varepsilon}^{i-1}\mathcal{R}_{n},\overline{\mathcal{K}}_{\varepsilon}^{j-1}\mathcal{R}_{n}\rangle_{\mathbb{K}},\label{EQ37}\\&=\langle\mathcal{H}_{m\times n}\left(i\right),\mathcal{H}_{m\times n}\left(j\right)\rangle_{\mathbb{K}},\label{EQ38}
\end{align}
where we mark $\overline{\mathcal{K}}_{\varepsilon}^{i-1}\mathcal{R}_{n}=\left[\overline{\mathcal{K}}_{\varepsilon}^{i-1}\mathbf{g}\left(\gamma_{0}\right),\ldots,\overline{\mathcal{K}}_{\varepsilon}^{i-1}\mathbf{g}\left(\gamma_{n\varepsilon}\right)\right]$ for convenience. Eq. (\ref{EQ38}) is derived from the fact that $\mathcal{H}_{m\times n}$ serves as the sampling of $\mathcal{R}_{n}$. Meanwhile, the left side of Eq. (\ref{EQ35}) coincides with the empirical Gramian matrix $\widehat{\mathcal{V}}$ associated with matrix $\mathcal{H}_{m\times n}$
\begin{align}
\widehat{\mathcal{V}}_{i,j}=\frac{1}{m}\langle\mathcal{H}_{m\times n}\left(i\right),\mathcal{H}_{m\times n}\left(j\right)\rangle.\label{EQ39}
\end{align}

Our formal proof can be developed based on these two properties. Let us consider the first $r<n$ element of $\mathcal{R}_{n}$ 
\begin{align}
\mathcal{R}_{r}&=\left[\mathbf{g}\left(\gamma_{0}\right),\overline{\mathcal{K}}_{\varepsilon}\mathbf{g}\left(\gamma_{0}\right),\overline{\mathcal{K}}_{\varepsilon}^{2}\mathbf{g}\left(\gamma_{0}\right),\ldots,\overline{\mathcal{K}}_{\varepsilon}^{r-1}\mathbf{g}\left(\gamma_{0}\right)\right],\label{EQ40}
\end{align}
which defines a possible basis of $\mathbb{K}$. 

Theoretically, the learned Koopman operator restricted to $\mathbb{K}$ can be represented by a companion matrix
\begin{align}
\mathcal{C}=\begin{bmatrix} 
	0 & 0 & \cdots & 0 & c_{0} \\
        1 & 0 & \cdots & 0 & c_{1} \\
        0 & 1 & \cdots & 0 & c_{2} \\
	\vdots & \vdots & \ddots & \vdots & \vdots\\
	0 & 0 & \cdots & 1 & c_{r-1} \\
	\end{bmatrix}.\label{EQ41}
\end{align}
The last column of $\mathcal{C}$ denotes the coordinate of $\overline{\mathcal{K}}_{\varepsilon}^{r}\mathbf{g}\left(\gamma_{0}\right)$ defined by the basis, which should be calculated as
\begin{align}
\mathcal{C}\left(r\right)=\mathcal{V}^{-1}\begin{bmatrix} 
	 \langle\mathbf{g}\left(\gamma_{0}\right),\overline{\mathcal{K}}_{\varepsilon}^{r}\mathbf{g}\left(\gamma_{0}\right)\rangle_{\mathbb{K}} \\ \langle\overline{\mathcal{K}}_{\varepsilon}\mathbf{g}\left(\gamma_{0}\right),\overline{\mathcal{K}}_{\varepsilon}^{r}\mathbf{g}\left(\gamma_{0}\right)\rangle_{\mathbb{K}} \\
       \langle\overline{\mathcal{K}}_{\varepsilon}^{2}\mathbf{g}\left(\gamma_{0}\right),\overline{\mathcal{K}}_{\varepsilon}^{r}\mathbf{g}\left(\gamma_{0}\right)\rangle_{\mathbb{K}} \\
	 \vdots\\
  \langle\overline{\mathcal{K}}_{\varepsilon}^{r-1}\mathbf{g}\left(\gamma_{0}\right),\overline{\mathcal{K}}_{\varepsilon}^{r}\mathbf{g}\left(\gamma_{0}\right)\rangle_{\mathbb{K}} \\
	\end{bmatrix}.\label{EQ42}
\end{align}

Empirically, the learned Koopman operator restricted to $\operatorname{span}\left(\mathcal{H}_{\left(m,n\right)}\right)$ can be also represented by a companion matrix, whose last column can be calculated as
\begin{align}
\widehat{\mathcal{C}}\left(r\right)= \frac{1}{m}\widehat{\mathcal{V}}^{-1}\begin{bmatrix} 
	 \langle\mathcal{H}_{m\times n}\left(1\right),\overline{\mathcal{K}}_{\varepsilon}^{r}\mathcal{H}_{m\times n}\left(1\right)\rangle \\ \langle\overline{\mathcal{K}}_{\varepsilon}\mathcal{H}_{m\times n}\left(1\right),\overline{\mathcal{K}}_{\varepsilon}^{r}\mathcal{H}_{m\times n}\left(1\right)\rangle \\
       \langle\overline{\mathcal{K}}_{\varepsilon}^{2}\mathcal{H}_{m\times n}\left(1\right),\overline{\mathcal{K}}_{\varepsilon}^{r}\mathcal{H}_{m\times n}\left(1\right)\rangle \\
	 \vdots\\
  \langle\overline{\mathcal{K}}_{\varepsilon}^{r-1}\mathbf{g}\left(\gamma_{0}\right),\overline{\mathcal{K}}_{\varepsilon}^{r}\mathbf{g}\left(\gamma_{0}\right)\rangle_{\mathbb{K}} \\
	\end{bmatrix}.\label{EQ43}
\end{align}

It is trivial to prove 
\begin{align}
\lim_{m\rightarrow\infty}\widehat{\mathcal{V}}^{-1}=\left(\lim_{m\rightarrow\infty}\widehat{\mathcal{V}}\right)^{-1}=\mathcal{V}^{-1} \label{EQ44}
\end{align}
applying Eq. (\ref{EQ36}) and Eqs. (\ref{EQ38}-\ref{EQ39}). Similarly, we can know 
\begin{align}
&\lim_{m\rightarrow\infty} \frac{1}{m}\langle\overline{\mathcal{K}}_{\varepsilon}^{k}\mathcal{H}_{m\times n}\left(1\right),\overline{\mathcal{K}}_{\varepsilon}^{r}\mathcal{H}_{m\times n}\left(1\right)\rangle_{\mathbb{K}}\notag\\=&\langle\overline{\mathcal{K}}_{\varepsilon}^{k}\mathbf{g}\left(\gamma_{0}\right),\overline{\mathcal{K}}_{\varepsilon}^{r}\mathbf{g}\left(\gamma_{0}\right)\rangle_{\mathbb{K}},\;\forall k\in\{0,\ldots,r-1\}.\label{EQ45}
\end{align}
Therefore, we can derive
\begin{align}
\lim_{m\rightarrow\infty}\widehat{\mathcal{C}}_{ij}=\mathcal{C}_{ij},\;\forall i,j\in\{1,\ldots,r\}^{2} \label{EQ46}
\end{align}
based on Eqs. (\ref{EQ44}-\ref{EQ45}), implying that
\begin{align}
\lim_{m\rightarrow\infty}\sum_{M\in \mathcal{PM}_{k}\left(\widehat{\mathcal{C}}\right)}M=\sum_{M\in \mathcal{PM}_{k}\left(\mathcal{C}\right)}M,\;\forall k\in\{1,\ldots,r\}. \label{EQ47}
\end{align}
In Eq. (\ref{EQ47}), notion $\mathcal{PM}_{k}\left(\cdot\right)$ denotes the set of all the $k$-order principal minors of the corresponding matrix. Now, let us consider the characteristic polynomials of $\widehat{\mathcal{C}}$ and $\mathcal{C}$
\begin{align}
P_{\widehat{\mathcal{C}}}\left(z\right)&=z^{r}+\sum_{i=1}^{r}\left(-1\right)^{i}\left(\sum_{M\in \mathcal{PM}_{i}\left(\widehat{\mathcal{C}}\right)}M\right)z^{r-i}, \label{EQ48}\\
P_{\mathcal{C}}\left(z\right)&=z^{r}+\sum_{i=1}^{r}\left(-1\right)^{i}\left(\sum_{M\in \mathcal{PM}_{i}\left(\mathcal{C}\right)}M\right)z^{r-i}, \label{EQ49}
\end{align}
whose distance at the limit of $m\rightarrow\infty$ can be measured as 
\begin{align}
&\lim_{m\rightarrow\infty}\Vert P_{\widehat{\mathcal{C}}}\left(z\right)-P_{\mathcal{C}}\left(z\right)\Vert\notag\\=&\lim_{m\rightarrow\infty}\max_{1\leq i\leq r}\Bigg\vert \left(-1\right)^{i}\left(\sum_{M\in \mathcal{PM}_{i}\left(\widehat{\mathcal{C}}\right)}M-\sum_{M\in \mathcal{PM}_{i}\left(\mathcal{C}\right)}M\right)\Bigg\vert, \label{EQ50}\\
=&0.\label{EQ51}
\end{align}
Because the roots of a given polynomial evolve continuously as the function of the coefficients, we know that $\widehat{\mathcal{C}}$ and $\mathcal{C}$ share the same set of eigenvalues at the limit of $m\rightarrow\infty$ since their characteristic polynomials converge to the same. Moreover, the convergence of $\widehat{\mathcal{C}}$ to $\mathcal{C}$ and the convergence of the eigenvalues $\widehat{\mathcal{C}}$ to those of $\mathcal{C}$ eventually imply the convergence of the eigenfunctions.

\begin{figure*}[t!]
\includegraphics[width=1\columnwidth]{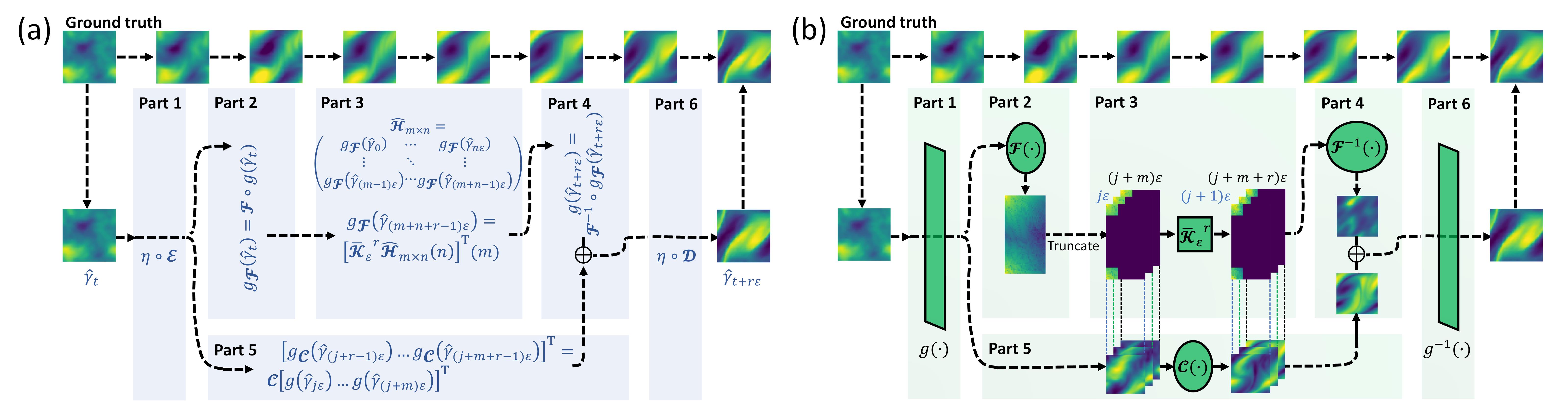}
\caption{\label{G1} Conceptual illustrations of neural network architectures of the original Koopman neural operator. (a) summarizes key mathematical transform in each part, where $r$ is the prediction length. (b) visualizes a prediction instance on the 2-dimensional Navier-Stokes equation.} 
 \end{figure*}

\subsection{The original Koopman neural operator: Computation}
In Ref. \cite{anonymous2023koopman}, we have proposed an architecture to implement the original Koopman neural operator on neural networks. The details of architecture designs are presented below:
\begin{itemize}
    \item \textbf{Part 1: Observation.} An encoder (e.g., a single non-linear layer with $\tanh\left(\cdot\right)$ activation function in the original Koopman neural operator) serves as observation function $\mathbf{g}\left(\cdot\right)$ to transform $\phi_{t}=\phi\left(D\times\{t\}\right)$, an arbitrary input of the PDE family (e.g., $\phi_{t}$ can be directly set as $\gamma_{t}$), into $\mathbf{g}\left(\widehat{\gamma_{t}}\right)\in\mathcal{G}\left(\mathbb{R}^{d_{\widehat{\gamma}}}\times T\right)$
    \begin{align}
\mathbf{g}\left(\widehat{\gamma_{t}}\right)=\operatorname{Encoder}\left(\phi_{t}\right),\;\forall t\in T.\label{EQ52}
\end{align}
Please see \textbf{Fig. 1} for illustrations.
    \item \textbf{Part 2: Fourier transform.} Similar to the Fourier neural operator \cite{li2020fourier}, the original Koopman neural operator also utilizes the Fourier transform during the iterative update of the Green function in Eq. (\ref{EQ5}). Given $\mathbf{g}\left(\widehat{\gamma_{t}}\right)$, we derive the Fourier transform $\mathbf{g}_{\mathcal{F}}\left(\cdot\right)=\mathcal{F}\circ\mathbf{g}\left(\cdot\right)$, where we truncate the Fourier series at $\omega$, a maximum frequency
     \begin{align}
\mathbf{g}_{\mathcal{F}}\left(\xi\right)&=\chi_{\left[0,\omega\right]}\left(\xi\right)\int_{D\times\{t\}}\mathbf{g}\left(\widehat{\gamma}\left(x_{t}\right)\right)\exp\left(-2\pi i\langle x_{t},\xi\rangle\right)\mathsf{d}x_{t}.\label{EQ53}
    \end{align}
    Note that $\chi_{\cdot}\left(\cdot\right)$ denotes the indicator function (i.e., $\chi_{A}\left(a\right)=1$ if $a\in A$, otherwise $\chi_{A}\left(a\right)=0$). Computationally, the above transform is implemented by fast Fourier transform. For convenience, we mark
    \begin{align}
\mathbf{g}_{\mathcal{F}}\left(\widehat{\gamma_{t}}\right)=\mathcal{F}\circ\mathbf{g}\left(\widehat{\gamma_{t}}\right)=\left\{\mathbf{g}_{\mathcal{F}}\left(\xi\right)\vert \xi\in\left[0,\infty\right)\right\}\label{EQ54}
    \end{align}
    as the transformed result of $\widehat{\gamma_{t}}$. Different from Ref. \cite{li2020fourier}, our main motivation for using the truncated Fourier transform is to extract the low-frequency information (i.e., main system components) of the represented equation solution $\mathbf{g}\left(\widehat{\gamma_{t}}\right)$. Certainly, frequency truncation inevitably causes the loss of high-frequency information (i.e., high-frequency perturbations or edges). In the original Koopman neural operator, \textbf{Part 5} is designed to complement the lost information associated with high-frequency. See \textbf{Fig. 1} for more details.
   \item \textbf{Part 3: Hankel representation and offline Koopman operator.} Once we have derived $\mathbf{g}_{\mathcal{F}}\left(\widehat{\gamma_{t}}\right)$ for every $t\in \varepsilon\mathbb{N}^{+}$, a Hankel matrix $\widehat{\mathcal{H}}_{m\times n}$ of $\mathbf{g}_{\mathcal{F}}\left(\widehat{\gamma_{t}}\right)$ will be generated following $m\in\mathbb{N}$, a dimension of delay-embedding (note that $n\in\mathbb{N}$ is the number of all accessible samples)
   \begin{align}
\widehat{\mathcal{H}}_{m\times n}&=\begin{bmatrix} 
	\mathbf{g}_{\mathcal{F}}\left(\widehat{\gamma}_{0}\right) & \mathbf{g}_{\mathcal{F}}\left(\widehat{\gamma}_{\varepsilon}\right) & \cdots & \mathbf{g}_{\mathcal{F}}\left(\widehat{\gamma}_{n\varepsilon}\right)  \\
	\vdots & \vdots & \vdots & \vdots\\
	\mathbf{g}_{\mathcal{F}}\left(\widehat{\gamma}_{\left(m-1\right)\varepsilon}\right)  & \mathbf{g}_{\mathcal{F}}\left(\widehat{\gamma}_{m\varepsilon}\right) & \cdots & \mathbf{g}_{\mathcal{F}}\left(\widehat{\gamma}_{\left(m+n-1\right)\varepsilon}\right) \\
	\end{bmatrix}.\label{EQ55}
    \end{align}
We train a $o\times o$ linear layer to learn a neural network representation of Koopman operator $\overline{\mathcal{K}}_{\varepsilon}:\mathcal{G}\left(\mathbb{R}^{d_{\widehat{\gamma}}}\times T\right)\rightarrow\widehat{\mathbb{K}}$ following Eqs. (\ref{EQ32}-\ref{EQ33}), where $\widehat{\mathbb{K}}$ is spanned by $\widehat{\mathcal{H}}_{m\times n}$. The learned $\overline{\mathcal{K}}_{\varepsilon}$ can be used to predict the future state of $\mathbf{g}_{\mathcal{F}}\left(\widehat{\gamma}_{\left(m+n-1\right)\varepsilon}\right)$ as 
\begin{align}
\mathbf{g}_{\mathcal{F}}\left(\widehat{\gamma}_{\left(m+n+r-1\right)\varepsilon}\right)=\left[\overline{\mathcal{K}}_{\varepsilon}^{r}\widehat{\mathcal{H}}_{m\times n}\left(n\right)\right]^{\mathsf{T}}\left(m\right),\;r\in\mathbb{N}^{+}\label{EQ56}
\end{align}
where notion $\mathsf{T}$ denotes the transpose of a matrix. Please see \textbf{Fig. 1} for illustrations.
   \item \textbf{Part 4: Inverse Fourier transform.} After $\mathbf{g}_{\mathcal{F}}\left(\widehat{\gamma}_{\left(m+n\right)\varepsilon}\right)$ is predicted in \textbf{Part 3}, it is transformed from the Fourier space to $\mathcal{G}\left(\mathbb{R}^{d_{\widehat{\gamma}}}\times T\right)$ by an inverse Fourier transform
   \begin{align}
\mathbf{g}\left(\widehat{\gamma}\left(x_{t}\right)\right)&=\frac{1}{\left(2\pi\right)^{d_{\widehat{\gamma}}}}\int_{-\infty}^{\infty}\mathbf{g}_{\mathcal{F}}\left(\xi\right)\exp\left(2\pi i\langle x_{t},\xi\rangle\right)\mathsf{d}\xi,\label{EQ57}
    \end{align}
where $t=\left(m+n+r-1\right)\varepsilon$. For convenience, we mark
\begin{align}
\mathbf{g}\left(\widehat{\gamma}_{\left(m+n+r-1\right)\varepsilon}\right)=\mathcal{F}^{-1}\circ\mathbf{g}_{\mathcal{F}}\left(\widehat{\gamma}_{\left(m+n+r-1\right)\varepsilon}\right).\label{EQ58}
\end{align}
Please see \textbf{Fig. 1} for instances of \textbf{Part 4}.
\item \textbf{Part 5: High-frequency information complement.} In the original Koopman neural operator, we use a convolutional layer to extract high-frequency of $\mathbf{g}\left(\widehat{\gamma_{t}}\right)$ because convolutional layers can amplify high-frequency components according to Fourier analysis \cite{park2022vision}. Therefore, we train a convolutional layer $\mathcal{C}$ on the outputs of \textbf{Part 1} to extract their high-frequency information. As a complement of \textbf{Parts 2-4}, the convolutional layer realizes a forward prediction of high-frequency information
\begin{align} &\left[\mathbf{g}_{\mathcal{C}}\left(\widehat{\gamma}_{\left(j+r-1\right)\varepsilon}\right),\ldots,\mathbf{g}_{\mathcal{C}}\left(\widehat{\gamma}_{\left(j+m+r-1\right)\varepsilon}\right)\right]^{\mathsf{T}}\notag\\=&\mathcal{C}\left[\mathbf{g}\left(\widehat{\gamma}_{j\varepsilon}\right),\ldots,\mathbf{g}\left(\widehat{\gamma}_{\left(j+m\right)\varepsilon}\right)\right]^{\mathsf{T}},\;\forall j=1,\ldots,n.\label{EQ59}
\end{align}
See \textbf{Fig. 1} for illustrations.
   \item \textbf{Part 6: Inverse observation.} Once two future states, $\mathbf{g}\left(\widehat{\gamma}_{\left(m+n\right)\varepsilon}\right)$ and $\mathbf{g}_{\mathcal{C}}\left(\widehat{\gamma}_{\left(m+n\right)\varepsilon}\right)$, are predicted by \textbf{Parts 2-4} and \textbf{Part 5}, they are unified in a linear manner
   \begin{align}
\mathbf{g}_{\mathcal{U}}\left(\widehat{\gamma}_{\left(m+n+r-1\right)\varepsilon}\right)=\mathbf{g}\left(\widehat{\gamma}_{\left(m+n+r-1\right)\varepsilon}\right)+\mathbf{g}_{\mathcal{C}}\left(\widehat{\gamma}_{\left(m+n+r-1\right)\varepsilon}\right).\label{EQ60}
   \end{align}
   Given $\mathbf{g}_{\mathcal{U}}\left(\widehat{\gamma}_{\left(m+n\right)\varepsilon}\right)$, a non-linear decoder (e.g., a single non-linear layer with $\tanh\left(\cdot\right)$ activation function in the original neural operator) is trained to approximate the inverse of observation function 
   \begin{align}
\mathbf{g}^{-1}\left(\cdot\right)\simeq\mathbf{g}_{\mathcal{U}}^{-1}\left(\cdot\right)\label{EQ61}
   \end{align}
   and derive 
   \begin{align}
\widehat{\gamma}_{\left(m+n+r-1\right)\varepsilon}=\operatorname{Decoder}\left(\mathbf{g}_{\mathcal{U}}\left(\widehat{\gamma}_{\left(m+n+r-1\right)\varepsilon}\right)\right)\label{EQ62}
   \end{align}
   as the target state of equation solution in space $\mathbb{R}^{d_{\widehat{\gamma}}}$. Please see \textbf{Fig. 1} for illustrations.
\end{itemize}

\textbf{Parts 1-6} define the iterative update strategy of Eq. (\ref{EQ23}). For any $t^{\prime}>t\in\varepsilon\mathbb{N}$, the iterative dynamics is given as
\begin{align}
\widehat{\gamma}_{t^{\prime}}=&\Big[\mathbf{g}^{-1}\Big(\underbrace{\mathcal{F}^{-1}\circ\overline{\mathcal{K}}_{\varepsilon}^{t^{\prime}-t}\circ\mathcal{F}\circ\mathbf{g}\left(\widehat{\gamma}_{\left[t-m\varepsilon,t\right]}\right)}_{\textbf{Parts 1-4}}\notag\\&+\underbrace{\mathcal{C}\circ\mathbf{g}\left(\widehat{\gamma}_{\left[t-m\varepsilon,t\right]}\right)}_{\textbf{Part 1 and part 5}}\Big)\Big]^{\mathsf{T}}\left(m\right),\label{EQ63}
\end{align}
in which $\widehat{\gamma}_{\left[t-m\varepsilon,t\right]}$ denotes a vector $\left[\widehat{\gamma}_{t-m\varepsilon},\ldots,\widehat{\gamma}_{t}\right]$. The loss function for optimizing Eq. (\ref{EQ63}) is defined as
\begin{align}
\mathcal{L}=\lambda_{p}\Vert \widehat{\gamma}_{t^{\prime}}-\gamma_{t^{\prime}}\Vert_{F}+\lambda_{r}\sum_{i=0}^{m}\Vert\mathbf{g}^{-1}\circ\mathbf{g}\left(\widehat{\gamma}_{t-im\varepsilon}\right)-\gamma_{t-im\varepsilon}\Vert_{F},\label{EQ64}
\end{align}
where $\lambda_{p},\lambda_{r}\in\left(0,\infty\right)$ denotes the weights of prediction and reconstruction processes in the loss function.  

The one-unit architecture of the original Koopman neural operator is visualized in \textbf{Fig. 1}. A multi-unit architecture can be readily constructed by cascading the copy of \textbf{Parts 2-5} multiple times. 

\section{The Koopman neural operator family}
Beyond the original Koopman neural operator (KNO) \cite{anonymous2023koopman}, we generalize it to four kinds of variants to fit in with different application demands.

\subsection{The compact KNO sub-family: Definition}
The compact KNO sub-family includes two compact variants of KNO realized by
multi-layer perceptrons (MLP-KNO) and convolutional neural networks (CNN-KNO). These variants are proposed to
accurately solve PDEs with small model sizes. Specifically, they are designed following Eqs. (\ref{EQ52}-\ref{EQ62}), where encoder and decoder are defined as 
\begin{align}
   \operatorname{Encoder}&=\eta\circ\mathcal{E},\;\mathcal{E}\in\{\mathcal{W}_{e},\mathcal{C}_{e}\},\label{EQ65}\\
  \operatorname{Decoder}&=\eta\circ \mathcal{D},\;\mathcal{D}\in\{\mathcal{W}_{d},\mathcal{C}_{d}\}.\label{EQ66}
    \end{align}
In Eqs. (\ref{EQ65}-\ref{EQ66}), mapping $\eta$ denotes a non-linear activation function (e.g., we use $\tanh\left(\cdot\right)$ in our research), notions $\mathcal{W}_{e}$ and $\mathcal{W}_{d}$ are two weight matrices of the corresponding sizes, and $\mathcal{C}_{e}$ and $\mathcal{C}_{d}$ are two convolutional layers. 

 Our proposed KoopmanLab module offers user-friendly tools to customize an MLP-KNO. A customized instance is presented below.
 \begin{lstlisting}[language=Python]
import koopmanlab as kp

MLP_KNO1D = model.koopman(backbone = "KNO1d", autoencoder = "MLP", o = o, f = f, r = r, device = device)

MLP_KNO2D = model.koopman(backbone = "KNO2d", autoencoder = "MLP", o = o, f = f, r = r, device = device)

MLP_KNO1D.compile()
MLP_KNO2D.compile()

## Parameter definitions:
#  o: the dimension of the learned Koopman operator
#  f: the number of frequency modes below frequency truncation threshold
#  r: the power of the Koopman operator in EQ. (56)
#  device: if CPU or GPU is used for computation
\end{lstlisting}

Similarly, a CNN-KNO can be customized using the following code, where we present 1-dimensional and 2-dimensional visions.
 \begin{lstlisting}[language=Python]
import koopmanlab as kp

CNN_KNO_1D = model.koopman(backbone = "KNO1d", autoencoder = "Conv1d", o = o, f = f, r = r, device = device)

CNN_KNO_1D.compile()

CNN_KNO_2D = model.koopman(backbone = "KNO2d", autoencoder = "Conv2d", o = o, f = f, r = r, device = device)

CNN_KNO_1D.compile()
CNN_KNO_2D.compile()

## Parameter definitions:
#  o: the dimension of the learned Koopman operator
#  f: the number of frequency modes below frequency truncation threshold
#  r: the power of the Koopman operator in EQ. (56)
#  device: if CPU or GPU is used for computation
\end{lstlisting}

\subsection{The compact KNO sub-family: Validation}
To validate the proposed compact KNO sub-family in PDE solving tasks, we design mesh-independent and long-term prediction tasks on representative PDEs (e.g., the 2-dimensional Navier-Stokes equation \cite{wang1991exact} and the 1-dimensional Bateman–Burgers equation \cite{benton1972table}). The numerical data sets of these two are provided by Ref. \cite{li2020fourier}.

\begin{figure*}[t!]
\includegraphics[width=1\columnwidth]{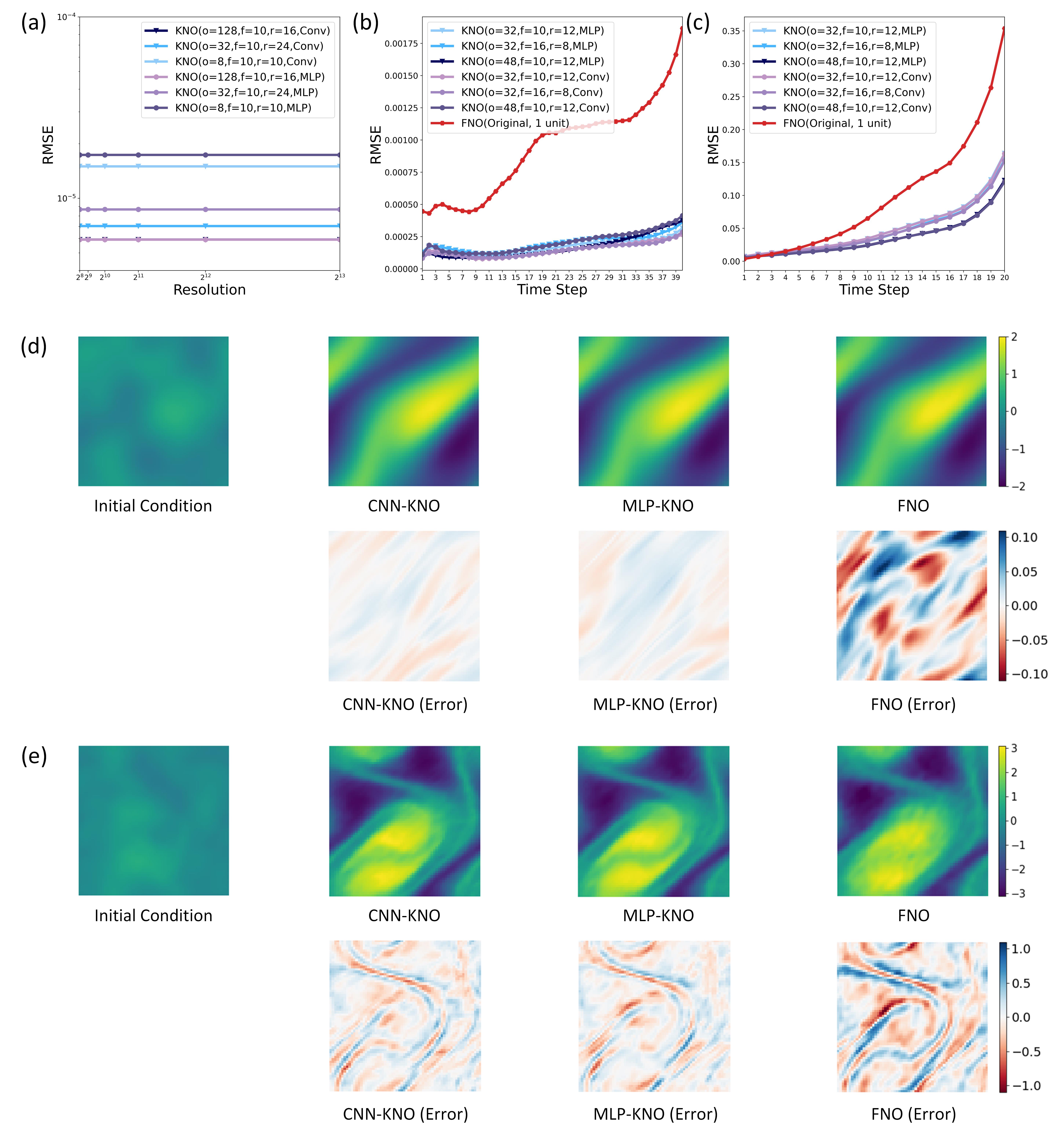}
\caption{\label{G2} Experimental validation of the compact KNO sub-family. (a) Results of the mesh-independent experiment on the Bateman–Burgers equation. (b) Results of the long-term prediction experiment on the 2-dimensional Navier-Stokes equation with a viscosity coefficient $\nu=10^{-3}$. (b) Results of the same long-term prediction experiment on the 2-dimensional Navier-Stokes equation with $\nu=10^{-4}$. (d) The prediction results and errors (RMSE) on the 2-dimensional Navier-Stokes equation with $\nu=10^{-3}$. (e) The prediction results and errors (RMSE) on the 2-dimensional Navier-Stokes equation with $\nu=10^{-4}$.} 
 \end{figure*}

 \begin{table*}[!t]
\centering
\small{
\begin{tabular}{l|l|l}
 \hline
Models   & Settings   & Parameter number                                    \\ 
\hline
FNO & default settings, one-unit \cite{li2020fourier} & 233897       \\
MLP-KNO & $\left(o,f,r\right)=\left(32,10,12\right)$  & 206538       \\  
MLP-KNO & $\left(o,f,r\right)=\left(32,16,8\right)$  & 526026       \\    
MLP-KNO & $\left(o,f,r\right)=\left(48,10,12\right)$  & 464170      \\  
CNN-KNO & $\left(o,f,r\right)=\left(32,10,12\right)$  & 206538      \\  
CNN-KNO & $\left(o,f,r\right)=\left(32,16,8\right)$  & 526026      \\  
CNN-KNO & $\left(o,f,r\right)=\left(48,10,12\right)$  & 464170      \\  
 \hline
\end{tabular}
}
\caption{The parameter numbers of all implemented models in \textbf{Figs. 2(b-c)} counted by the tool provided by Ref. \cite{pathak2022fourcastnet,kurth2022fourcastnet}.}
\end{table*}

Specifically, the incompressible 2-dimensional Navier-Stokes equation has a vorticity form
\begin{align}
    \partial_{t} \gamma\left(x_{t}\right)+\chi\left(x_{t}\right)\nabla\gamma\left(x_{t}\right)&=\nu \Delta \gamma\left(x_{t}\right)+\psi\left(x_{t}\right),\notag\\&x_{t}\in\left(0,1\right)^{2}\times\left(0,\infty\right),\label{EQ67}\\
    \nabla \chi\left(x_{t}\right)&=0,\;x_{t}\in\left(0,1\right)^{2}\times\left(0,\infty\right),\label{EQ68}\\
\gamma\left(x_{0}\right)&=\gamma_{I},\;x_{0}\in\left(0,1\right)\times\{0\},\label{EQ69}
\end{align}
in which $\gamma\left(\cdot\right)$ denotes the vorticity, $\chi\left(\cdot\right)$ measures the velocity, $\psi\left(\cdot\right)$ is a time-independent forcing term. The viscosity coefficient is $\nu\in\{10^{-3},10^{-4}\}$ in our research. Given the data with the highest mesh resolution, one can further generate the data with the lower resolution by direct down-sampling \cite{li2020fourier}. The data with the highest mesh resolution has $2^{13}$ grids \cite{li2020fourier}. Our KoopmanLab module offers a function to load the data of the incompressible 2-dimensional Navier-Stokes equation.
 \begin{lstlisting}[language=Python]
import koopmanlab as kp

train_loader, test_loader = kp.data.navier_stokes(path, batch_size = 10, T_in = 10, T_out = 40, type = "1e-3", sub = 1)

## Parameter definitions:
#  path: the file path of the downloaded data set
#  T_in: the duration length of input data
#  T_out: the duration length required to predict
#  Type: the viscosity coefficient
#  sub: the down-sampling scaling factor. For instance, a scaling factor sub=2 acting on a 2-dimensional data with the spatial resolution 64*64 will create a down-sampled space of 32*32. The same factor action on a 1-dimensional data with the spatial resolution 1*64 implies a down-sampled space of 1*32
\end{lstlisting}

 The 1-dimensional Bateman–Burgers equation is defined as
\begin{align}
    \partial_{t} \gamma\left(x_{t}\right)+\partial_{x}\left(\frac{\gamma^{2}\left(x_{t}\right)}{2}\right)&=\nu \partial_{xx} \gamma\left(x_{t}\right),\;x_{t}\in\left(0,1\right)\times\left(0,1\right],\notag\\&x_{t}\in\left(0,1\right)\times\left(0,1\right],\label{EQ70}\\
\gamma\left(x_{0}\right)&=\gamma_{I},\;x_{0}\in\left(0,1\right)\times\{0\},\label{EQA71}
\end{align}
in which $\gamma_{I}$ stands for a periodic initial condition $\gamma_{I} \in L^{2}_{\text{periodic}}\left[\left(0,1\right);\mathbb{R}\right]$ and parameter $\nu\in\left(0,\infty\right)$ is the viscosity coefficient. We set $\nu=100$ in our research. The data with highest mesh resolution has $2^{16}$ grids \cite{li2020fourier}. To load this data set, one can use the following function.

 \begin{lstlisting}[language=Python]
import koopmanlab as kp

train_loader, test_loader = kp.data.burgers(path, batch_size = 64, sub = 32)

## Parameter definitions:
#  path: the file path of the downloaded data set
#  sub: the down-sampling scaling factor. For instance, a scaling factor sub=2 acting on a 2-dimensional data with the spatial resolution 64*64 will create a down-sampled space of 32*32. The same factor action on a 1-dimensional data with the spatial resolution 1*64 implies a down-sampled space of 1*32
\end{lstlisting}

In \textbf{Fig. 2(a)}, we validate the mesh-independent property of the proposed compact KNO sub-family adopting the same setting used in our earlier work \cite{anonymous2023koopman}. The mesh-independent property, as suggested by Refs. \cite{lu2019deeponet,bhattacharya2020model,nelsen2021random,li2020neural,li2020fourier,kovachki2021universal,li2022fourier}, arises from the fact that the neural operator is expected to learn the solution operator of an entire PDE family rather than be limited to a concrete parameterized instance. Specifically, we conduct the experiments on the data of 1-dimensional Bateman–Burgers equation associated with different mesh granularity conditions (i.e., spatial resolution of meshes). Different versions of the compact KNO sub-family are defined by changing hyper-parameters (e.g., operator size $o$, frequency mode number $f$, and the power of the Koopman operator $r=\frac{t^{\prime}-t}{\varepsilon}$ in Eq. (\ref{EQ56})). These models are trained by 1000 randomly selected samples with the lowest spatial resolution and conduct 1-second forward prediction on 200 samples associated with different resolutions. Batch size is set as $64$, the learning rate is initialized as $0.001$ and halved every 100 epochs, and the weights of prediction and reconstruction in Eq. (\ref{EQ64})) are set as $\left(\lambda_{p},\lambda_{r}\right)=\left(5,0.5\right)$. As shown in \textbf{Fig. 2(a)}, the prediction errors of all versions of the compact KNO sub-family maintain constantly across different spatial resolutions, suggesting the capacity of the compact KNO sub-family to be mesh-independent. Mesh-independence is important for PDE solving because it allows one to train a neural-network-based PDE solver on the data with low spatial resolution and directly apply the solver on the data with high spatial resolution, which breaks the trade-off of accuracy and efficiency in PDE solving. In our earlier work \cite{anonymous2023koopman}, one can further see a detailed comparison between the original KNO and FNO \cite{li2020fourier} in mesh-independent prediction task, where the original KNO outperforms FNO with a much smaller model size (e.g., a size of $5\times 10^{3}$ for KNO and a size of $2\times 10^{7}$ for FNO). Other neural operator models, such as graph neural operator (GNO) \cite{li2020neural} and multipole graph neural operator (MGNO) \cite{li2020multipole}, are no longer considered because they have been demonstrated as less accurate than FNO as reported by Ref. \cite{li2020fourier}.

In \textbf{Fig. 2(b-e)}, we validate the compact KNO sub-family by a long-term prediction task designed on the 2-dimensional Navier-Stokes equation data sets with viscosity coefficients $\nu=10^{-3}$ (\textbf{Fig. 2(b)}) and $\nu=10^{-4}$ (\textbf{Fig. 2(c)}). A down-sampling scaling factor of $2$ is defined to generate the data sets with $2^{12}$ grids. For comparison, a one-unit FNO is defined following the default setting introduced in Ref. \cite{li2020fourier}. A 40-time-interval prediction task is conducted on the data set with $\nu=10^{-3}$, where models are trained on 1000 samples of $\gamma\left(\left(0,1\right)^{2}\times\left[0,10\right)\right)$ and tested on 200 samples of $\gamma\left(\left(0,1\right)^{2}\times\left(10,50\right]\right)$. Similarly, a more challenging 10-time-interval prediction task is conducted on the data set with $\nu=10^{-4}$, in which models are trained on 8000 samples of $\gamma\left(\left(0,1\right)^{2}\times\left[0,10\right)\right)$ and tested on 200 samples of $\gamma\left(\left(0,1\right)^{2}\times\left(10,20\right]\right)$. \textbf{Fig. 2(b-c)} report the prediction performance of all models as the function of increasing prediction duration length. \textbf{Fig. 2(d-e)} visualize predicted instances and errors in the cases with $\nu=10^{-3}$ (\textbf{Fig. 2(d)}) and $\nu=10^{-4}$ (\textbf{Fig. 2(e)}). All experiment results suggest the optimal potential of the compact KNO sub-family in characterizing the long-term evolution of PDE solutions. Combining these results with the model sizes measured in \textbf{Table. 1}, we suggest that the compact KNO sub-family realizes a better balance between accuracy and efficiency because a KNO variant with a smaller model size can still outperform FNO significantly.

\subsection{The ViT-KNO sub-family: Definition}
Different from the compact KNO sub-family, the ViT-KNO sub-family is proposed for dealing with more intricate situations (here ViT stands for Vision Transformer \cite{dosovitskiy2020image}). Numerous applications of PDE solving (e.g., global climate forecasting) require the solver to be able to capture the underlying patterns of ultra-large data sets that may be related with certain unknown PDEs. Meanwhile, there may exist multiple variables of interest that govern by a group of coupled PDEs. To fit in with these situations, we follow the main idea of the compact KNO sub-family to develop a kind of transfer-based PDE solver. The mechanism underlying the proposed ViT-KNO sub-family is not completely same as Eqs. (\ref{EQ52}-\ref{EQ62}) because some mathematical details are modified to improve model applicability on noisy real data sets. We suggest the benefits of our modifications based on an experiment on ERA5, one of the largest data set of global atmospheric, land, and oceanic climate fields \cite{hersbach2020era5}. Nevertheless, more in-depth mathematical analyses of these modifications remain for future studies. 

Let us consider a case where there exist $v$ coupled variables, $\{\gamma^{1},\ldots,\gamma^{h}\}$, defined on domain $D\times T$. The dynamics of these variables are govern by a group of PDEs with unknown expressions. The objective is to learn the equation solutions of these latent PDEs such that the dynamics of $\{\gamma^{1},\ldots,\gamma^{h}\}$ can be accurately characterized.

The architecture of ViT-KNO sub-family consists of 7 parts. Below, we present a detailed computational implementation of each part.

\begin{itemize}
    \item \textbf{Part 1: Observation.} Similar to the encoder design in the compact KNO sub-family, an encoder component is implemented in ViT-KNO to serve as observation function $\mathbf{g}\left(\cdot\right)$ and transform $\phi_{t}^{s}=\phi^{s}\left(D\times\{t\}\right)$ into $\mathbf{g}\left(\widehat{\gamma}_{t}^{s}\right)$ for each $s\in\{1,\ldots,h\}$. Specifically, the encoder is realized by the token embedding layer in Vision Transformer (ViT) \cite{dosovitskiy2020image}. Given a joint input $\left[\phi_{t}^{1},\ldots,\phi_{t}^{h}\right]\in\mathbb{R}^{d_{\phi} \times h}$, we first transform it into a 3-dimensional token tensor $\Phi_{t}$ by a convolutional layer $\mathcal{C}_{e}$
    \begin{align}
\mathcal{C}_{e}\left(\left[\phi_{t}^{1},\ldots,\phi_{t}^{h}\right]\right)=\mathbf{g}\left(\widehat{\Gamma}_{t}\right)\in\mathbb{R}^{u\times v \times l},\label{EQ72}
\end{align}
where domain $D$ is reorganized into $u\times v$ patches (i.e., tokens). The patch is a kind of square and non-overlapping macro-mesh. If domain $D$ has been already discretized into multiple meshes, then the size of a patch equals the number of meshes it covers. Parameter $l$ denotes a customized embedding dimension, which is not necessarily the same as $h$. The derived tensor $\mathbf{g}\left(\widehat{\Gamma}_{t}\right)$ denotes the joint representation of $\left[\mathbf{g}\left(\widehat{\gamma}_{t}^{1}\right),\ldots,\mathbf{g}\left(\widehat{\gamma}_{t}^{l}\right)\right]$. Please see \textbf{Fig. 3} for illustrations.
    \item \textbf{Part 2: Fourier transform.} Similar to adaptive Fourier neural operator \cite{pathak2022fourcastnet,kurth2022fourcastnet,guibas2021adaptive}, a truncated Fourier transform is applied on the first two dimensions of $\mathbf{g}\left(\widehat{\Gamma}_{t}\right)$ to derive the Fourier series of each embedded variable $s\in\{1,\ldots,l\}$
     \begin{align}
&\mathbf{g}_{\mathcal{F}}^{s}\left(\xi\right)\notag\\=&\chi_{\left[0,\omega\right]}\left(\xi\right)\int_{\left[u\right]\times\left[v\right]\times\{t\}}\mathbf{g}\left(\widehat{\gamma}^{s}\left(x_{t}\right)\right)\exp\left(-2\pi i\langle x_{t},\xi\rangle\right)\mathsf{d}x_{t},\label{EQ73}
    \end{align}
    where $\left[u\right]=\{1,\ldots,u\}$. For convenience, we mark
    \begin{align}
\mathbf{g}_{\mathcal{F}}^{s}\left(\widehat{\gamma_{t}}\right)=\mathcal{F}\circ\mathbf{g}\left(\widehat{\gamma}_{t}^{s}\right)=\left\{\mathbf{g}_{\mathcal{F}}^{s}\left(\xi\right)\vert \xi\in\left[0,\infty\right)\right\},\label{EQ74}\\
\mathbf{g}_{\mathcal{F}}\left(\widehat{\Gamma}_{t}\right)=\mathcal{F}\circ\mathbf{g}\left(\widehat{\Gamma}_{t}\right)=\left[\mathbf{g}_{\mathcal{F}}^{1}\left(\widehat{\gamma_{t}}\right),\ldots,\mathbf{g}_{\mathcal{F}}^{l}\left(\widehat{\gamma_{t}}\right)\right],\label{EQ75}
    \end{align}
    in which $s\in\{1,\ldots,l\}$.
    Similar to the compact KNO sub-family, frequency truncation leads to the loss of high-frequency information. In the ViT-KNO sub-family, \textbf{Part 5} is designed for complementing high-frequency information. See \textbf{Fig. 3} for details.
 
   \item \textbf{Part 3: Koopman-operator-associated component.} After deriving $\mathbf{g}_{\mathcal{F}}\left(\widehat{\Gamma}_{t}\right)$ for every $t\in \varepsilon\mathbb{N}^{+}$, a Koopman-operator-associated component is designed to function on the third dimension of every token in $\mathbf{g}_{\mathcal{F}}\left(\widehat{\Gamma}_{t}\right)$ and realize the iterative dynamics
   \begin{align}
\mathbf{g}_{\mathcal{F}}\left(\widehat{\Gamma}_{t+\varepsilon}\right)&=\overline{\mathcal{K}}_{\varepsilon}\mathcal{S}\mathbf{g}_{\mathcal{F}}\left(\widehat{\Gamma}_{t}\right)\notag\\&=\left[\mathbf{g}_{\mathcal{F}}^{1}\left(\widehat{\gamma}_{t+\varepsilon}\right),\ldots,\mathbf{g}_{\mathcal{F}}^{l}\left(\widehat{\gamma}_{t+\varepsilon}\right)\right],\label{EQ76}
    \end{align}
    in which Koopman operator $\overline{\mathcal{K}}_{\varepsilon}$ is learned by a linear transform and layer $\mathcal{S}$ is constructed by a non-linear activation function $\eta$ acting on a linear layer $\mathcal{W}$
    \begin{align}
\mathcal{S}&=\eta\circ\mathcal{W}.\label{EQ77}
    \end{align}
    Although $\mathcal{S}$ is not a part of the original Koopman neural operator \cite{anonymous2023koopman}, including it can efficiently enhance the capacity of this component to characterize intricate large-scale data. In KoopmanLab, the leaky rectified linear unit (Leaky ReLU) \cite{radford2015unsupervised} is suggested as a default choice of $\mathcal{S}$, which can also reduce to the ReLU function as a special case. Please see \textbf{Fig. 3} for illustrations of \textbf{Part 3}.

 \begin{figure*}[t!]
\includegraphics[width=1\columnwidth]{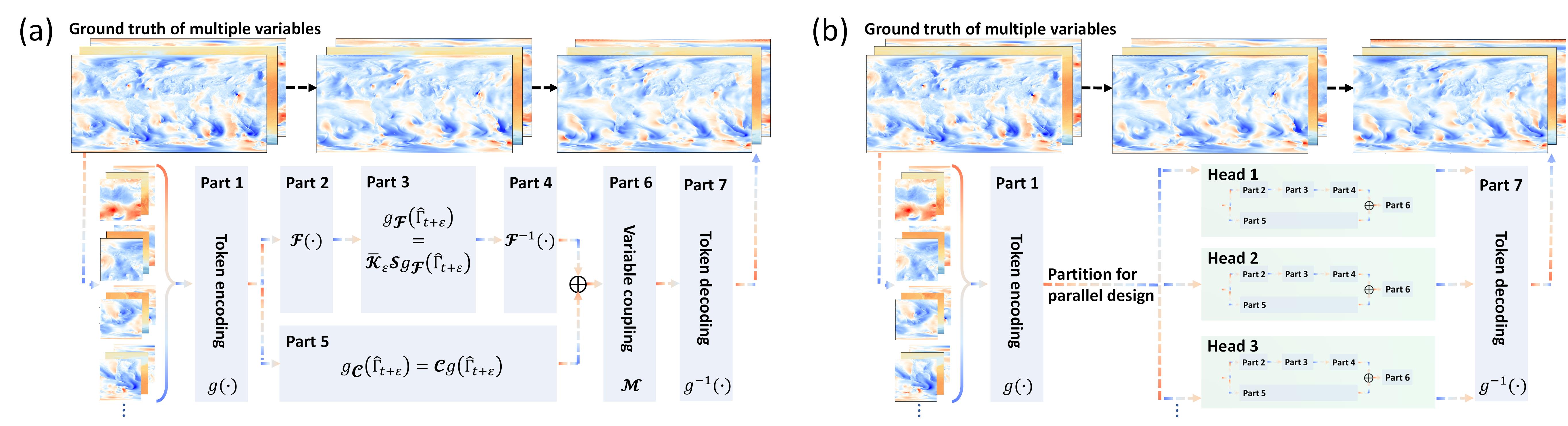}
\caption{\label{G3} Conceptual illustrations of neural network architectures of the ViT-KNO sub-family. (a) summarizes the computational design of each part. (b) illustrates an instance of the parallel design (i.e., multi-head design) of the ViT-KNO sub-family, where the depth of each head is 1.} 
 \end{figure*}
 
   \item \textbf{Part 4: Inverse Fourier transform.} Once $\mathbf{g}_{\mathcal{F}}\left(\widehat{\Gamma}_{t+\varepsilon}\right)$ is derived in \textbf{Part 3}, the inverse Fourier transform is applied on the first two dimensions of $\mathbf{g}_{\mathcal{F}}\left(\widehat{\Gamma}_{t+\varepsilon}\right)$ to transform $\mathbf{g}_{\mathcal{F}}\left(\widehat{\Gamma}_{t+\varepsilon}\right)$ back to the observation space
   \begin{align}
&\mathbf{g}\left(\widehat{\gamma}^{s}\left(x_{t+\varepsilon}\right)\right)\notag\\=&\frac{1}{\left(2\pi\right)^{d_{\widehat{\gamma}}}}\int_{-\infty}^{\infty}\mathbf{g}_{\mathcal{F}}^{s}\left(\xi\right)\exp\left(2\pi i\langle x_{t+\varepsilon},\xi\rangle\right)\mathsf{d}\xi,\label{EQ78}
    \end{align}
    based on which, we can define 
    \begin{align}
\mathbf{g}\left(\widehat{\Gamma}_{t+\varepsilon}\right)&=\mathcal{F}^{-1}\circ\mathbf{g}_{\mathcal{F}}\left(\widehat{\Gamma}_{t+\varepsilon}\right)\notag\\&=\left[\mathbf{g}\left(\widehat{\gamma}^{1}\left(x_{t+\varepsilon}\right)\right),\ldots,\mathbf{g}\left(\widehat{\gamma}^{l}\left(x_{t+\varepsilon}\right)\right)\right].\label{EQ79}
    \end{align}
Please see instances in \textbf{Fig. 3}.
\item \textbf{Part 5: High-frequency information complement.} Same as the compact KNO sub-family, there is a component for complementing high-frequency information in ViT-KNO. This component is also realized by a convolutional layer $\mathcal{C}$ that acts on the outputs of \textbf{Part 1} to learn the dynamics of high-frequency information
\begin{align} \mathbf{g}_{\mathcal{C}}\left(\widehat{\Gamma}_{t+\varepsilon}\right)=\mathcal{C}\mathbf{g}\left(\widehat{\Gamma}_{t}\right).\label{EQ80}
\end{align}
See \textbf{Fig. 1} for illustrations.
   \item \textbf{Part 6: Variable coupling.} Given two predicted states, $\mathbf{g}\left(\widehat{\Gamma}_{t+\varepsilon}\right)$ and $\mathbf{g}_{\mathcal{C}}\left(\widehat{\Gamma}_{t+\varepsilon}\right)$, by \textbf{Parts 2-4} and \textbf{Part 5}, we combine them in a linear form
   \begin{align}
\mathbf{g}_{\mathcal{U}}\left(\widehat{\Gamma}_{t+\varepsilon}\right)=\mathbf{g}\left(\widehat{\Gamma}_{t+\varepsilon}\right)+\mathbf{g}_{\mathcal{C}}\left(\widehat{\Gamma}_{t+\varepsilon}\right).\label{EQ81}
   \end{align}
   Because ViT-KNO is designed to learn multi-variate systems governed by unknown coupled PDEs, we need to characterize the coupling relation among variables. Because we lack the \emph{a priori} knowledge about these underlying PDEs, we suggest to capture the coupling relation by optimizing a non-linear layer $\mathcal{M}$
   \begin{align}
\mathbf{g}_{\mathcal{M}}\left(\widehat{\Gamma}_{t+\varepsilon}\right)&=\mathcal{M}\mathbf{g}_{\mathcal{U}}\left(\widehat{\Gamma}_{t+\varepsilon}\right).\label{EQ82}
   \end{align}
  Following the idea of adaptive Fourier neural operator \cite{pathak2022fourcastnet,kurth2022fourcastnet,guibas2021adaptive}, we use the Gaussian Error Linear Unit (GELU) as the activation function in this non-linear layer. Please see \textbf{Fig. 3} for illustrations.
  \item \textbf{Part 7: Inverse observation.} Given $\mathbf{g}_{\mathcal{M}}\left(\widehat{\Gamma}_{t+\varepsilon}\right)$, a decoder is implemented to function as the inverse of observation function 
   \begin{align}
\left[\widehat{\gamma}_{t+\varepsilon}^{1},\ldots,\widehat{\gamma}_{t+\varepsilon}^{h}\right]&\simeq\mathbf{g}^{-1}\left(\mathbf{g}_{\mathcal{M}}\left(\widehat{\Gamma}_{t+\varepsilon}\right)\right)\notag\\&=\operatorname{Decoder}\left(\mathbf{g}_{\mathcal{M}}\left(\widehat{\Gamma}_{t+\varepsilon}\right)\right).\label{EQ83}
   \end{align}
   Similar to the compact KNO sub-family, there are two kinds of decoders included in our proposed KoopmanLab module
   \begin{align}
\operatorname{Decoder}\in\{\mathcal{W}_{d},\mathcal{C}_{d}\},\label{EQ84}
    \end{align}
    in which $\mathcal{W}_{d}$ and $\mathcal{C}_{d}$ denote linear and convolutional layers, respectively. These two kinds of decoder designs distinguish between two variants of the ViT-KNO sub-family. See \textbf{Fig. 3} for illustrations.
\end{itemize}

\textbf{Parts 1-7} define the iterative update strategy of Eq. (\ref{EQ23}) in a multi-variate case. For any $t\in T$, the iterative dynamics is given as
\begin{align}
&\widehat{\Gamma}_{t+\varepsilon}\notag\\=&\mathbf{g}^{-1}\circ\mathcal{M}\circ\Big(\mathcal{F}^{-1}\circ\overline{\mathcal{K}}_{\varepsilon}\mathcal{S}\circ\mathcal{F}\circ\mathbf{g}\left(\widehat{\Gamma}_{t}\right)+\mathcal{C}\circ\mathbf{g}\left(\widehat{\Gamma}_{t}\right)\Big).\label{EQ85}
\end{align}
Multi-step prediction can be realized in an iterative manner. The loss function for optimizing Eq. (\ref{EQ85}) is 
\begin{align}
\mathcal{L}=\lambda_{p}\sum_{s=1}^{h}\Vert \widehat{\gamma}_{t+\varepsilon}^{s}-\gamma_{t+\varepsilon}^{s}\Vert_{F}+\lambda_{r}\sum_{s=1}^{h}\Vert\mathbf{g}^{-1}\circ\mathbf{g}\left(\widehat{\gamma}_{t}^{s}\right)-\gamma_{t}^{s}\Vert_{F},\label{EQ86}
\end{align}
where $\lambda_{p},\lambda_{r}\in\left(0,\infty\right)$ are the weights of prediction and reconstruction.  

Several computational tricks can be considered in the application. First, a LASSO regularization \cite{tibshirani1996regression} can be included to improve the robustness and sparsity of Koopman operator $\overline{\mathcal{K}}_{\varepsilon}$ in Eq. (\ref{EQ76}). This trick has been demonstrated as effective in adaptive Fourier neural operator \cite{pathak2022fourcastnet,kurth2022fourcastnet,guibas2021adaptive} and is applicable to the ViT-KNO sub-family as well. Second, the transformer architecture supports a parallel design the ViT-KNO sub-family. Specifically, the third dimension of the output of \textbf{Part 1} can be subdivided into multiple parts (e.g., $\mathbf{g}\left(\widehat{\Gamma}_{t}\right)\in\mathbb{R}^{u\times v\times l}$ is subdivided into $k$ parts such that each part is an element in $\mathbb{R}^{u\times v\times \frac{l}{k}}$). Then, \textbf{Parts 2-6} are copied $k\times j$ times, where each group of $j$ copies is organized into a sequence. Each sequence of $j$ copies is referred to as a head in the transformer, processes a corresponding $\frac{1}{k}$ part of $\mathbf{g}\left(\widehat{\Gamma}_{t}\right)\in\mathbb{R}^{u\times v\times l}$, and shares parameters during optimization (see \textbf{Fig. 3}). Computationally, parameters $k$ and $j$ are referred to as the number and the depth of heads. The processed outputs of these $k$ parallel heads are unified by \textbf{Part 7} to derive the final prediction result. In our proposed KoopmanLab, these two tricks are included to improve computational efficiency.

 Our KoopmanLab module supports customizing ViT-KNO frameworks. Below, we present an instance of ViT-KNO with a multi-layer perceptron as the decoder 
 \begin{lstlisting}[language=Python]
import koopmanlab as kp

ViT_KNO = model.koopman_vit(decoder = "MLP", resolution=(1440, 720), patch_size=(2, 2), in_chans=20, out_chans=20, head_num=20, embed_dim=768, depth = 16, parallel = True, high_freq = True, device=device)

ViT_KNO.compile()

## Parameter definitions:
#  resolution: the spatial resolution of input data
#  patch_size: the size of each patch (i.e., token)
#  in_chans: the number of target variables in the data set
#  out_chans: the number of predicted variables by ViT-KNO, which is usually same as in_chans
#  head_num: the number of heads
#  embed_dim: the embeding dimension denoted by l in Eq. (72)
#  depth: the depth of each head
#  parallel: if parallel design is applied
#  high_freq: if high-frequency information complement is applied
#  device: if CPU or GPU is used for computation
\end{lstlisting}

Similarly, a ViT-KNO whose decoder is a convolutional layer can be defined as the following
 \begin{lstlisting}[language=Python]
import koopmanlab as kp

ViT_KNO = model.koopman_vit(decoder = "Conv2d", resolution=(1440, 720), patch_size=(2, 2), in_chans=20, out_chans=20, head_num=20, embed_dim=768, depth = 16, parallel = True, high_freq = True, device=device)

ViT_KNO.compile()

## Parameter definitions:
#  resolution: the spatial resolution of input data
#  patch_size: the size of each patch (i.e., token)
#  in_chans: the number of target variables in the data set
#  out_chans: the number of predicted variables by ViT-KNO, which is usually same as in_chans
#  head_num: the number of heads
#  embed_dim: the embeding dimension denoted by l in Eq. (72)
#  depth: the depth of each head
#  parallel: if parallel design is applied
#  high_freq: if high-frequency information complement is applied
#  device: if CPU or GPU is used for computation
\end{lstlisting}

Please note that there exist some detailed model parameters that are not covered by the above codes because they are highly coupled during computation or less important in our theoretical derivations. Users are suggested to adjust them after loading the source code of ViT-KNO.

\subsection{The ViT-KNO sub-family: Validation}
To validate the proposed ViT-KNO sub-family, we implement a large-scale experiment on  ERA5, one of the largest high-
resolution data sets of global-scale multi-variate climate fields \cite{hersbach2020era5}. This data set has been extensively applied in weather forecasting tasks (e.g., see FourCastNet \cite{pathak2022fourcastnet,kurth2022fourcastnet}), ensuring the reproducibility and comparability of our results.

\begin{figure*}[t!]
\includegraphics[width=1\columnwidth]{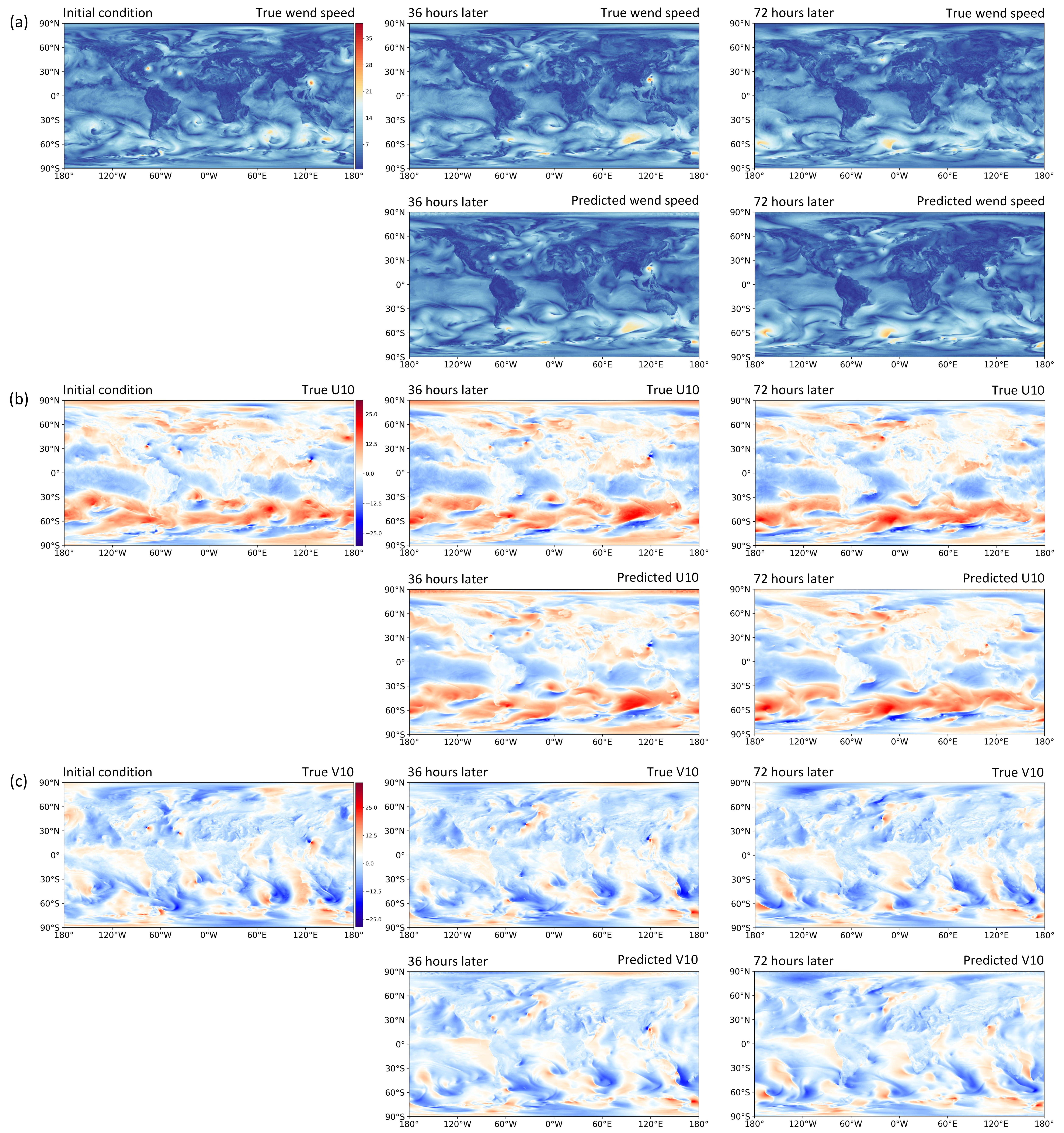}
\caption{\label{G4} Visualization of three instances of the experiment results on ERA5 data set. (a-c) respectively show the time-dependent predicted results of global wind speed, U10, and V10 variables on selected moments, accompanied by corresponding ground truths. Please note that wind speed is not originally included in the 20 climate variables selected from ERA5. It is calculated as $\text{wind speed}=\sqrt{\text{U10}^{2}+\text{V10}^{2}}$.} 
 \end{figure*}

  \begin{figure*}[t!]
\includegraphics[width=1\columnwidth]{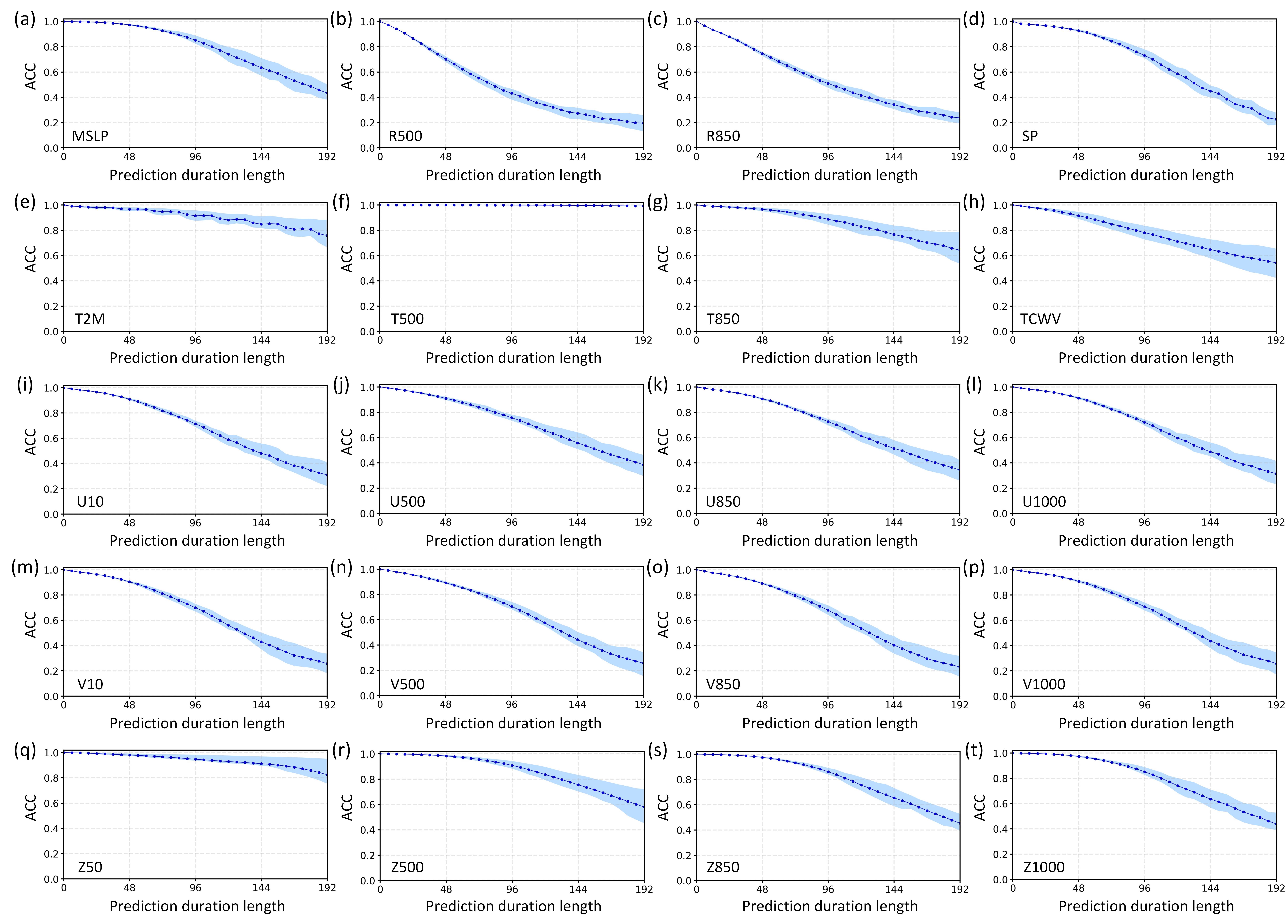}
\caption{\label{G5} Time-dependent prediction accuracy (measured by anomaly correlation coefficient, ACC) of ViT-KNO on 20 variables of ERA5 data set. The colored area denotes the interval of accuracy whose boundaries are fractiles. The dashed line denotes the average accuracy. } 
 \end{figure*}

Twenty important climate variables are considered in our research, including mean large-scale precipitation (MSLP), relative humidity with $500$ hPa (R500), relative humidity with $850$ hPa (R850), surface pressure (SP), 2m temperature (T2M), temperature with $500$ hPa (T500), temperature with $850$ hPa (T850), total column water vapour (TCWV), the 10m u-component of wind (U10), the u-component of wind with $500$ hPa (U500), the u-component of wind with $850$ hPa (U850), the u-component of wind with $1000$ hPa (U1000), the 10m v-component of wind (V10), the v-component of wind with $500$ hPa (V500), the v-component of wind with $850$ hPa (V850), the v-component of wind with $1000$ hPa (V1000), the geopotential with $50$ hPa (Z50), the geopotential with $500$ hPa (Z500), the geopotential with $850$ hPa (Z850), and the geopotential with $1000$ hPa (Z1000).

We test a ViT-KNO whose decoder is a multi-layer perceptron in a long-term prediction task. Given the samples of initial conditions, the ViT-KNO is required to predict the future states of all 20 climate variables after $t\in\{6,12,18,\ldots,192\}$ hours. The training data set includes the samples recorded from 1979 to 2015. The validation data set includes the samples recorded during 2016 and 2017. The test data set includes the samples recorded in 2018. The spatial resolution of all samples is set as $1440\times 720$. All data is pre-processed in a standard manner following Refs. \cite{hersbach2020era5,pathak2022fourcastnet,kurth2022fourcastnet}, where a $Z$-transform is applied to normalize all variables. The training of our ViT-KNO is implemented in a multi-GPU environment with $128\times 16$ GBs in total. The actual memory cost of training is $1250.56$ GBs. The testing of trained ViT-KNO is implemented in a single 16-GB GPU environment. 

Key model settings are summarized below. The batch size is set as $128$, the learning rate is $5\times 10^{-4}$ and updated by a cosine annealing approach, the patch size is set as $8\times 8$, the number of heads is $8$, the depth of heads is $12$, the embedded dimension $l$ in Eq. (\ref{EQ72}) is $768$, the relative weights of prediction and reconstruction in the loss function are $\lambda_{p}=0.9$ and $\lambda_{r}=0.1$, and the number of kept low-frequency modes after the fast Fourier transform is $32$. The defined model has $74691840$ parameters in total. All 20 climate variables are learned and predicted together rather than respectively. Please note that the information of land-form is not provided to the model. The model is required to learn climate variables with no additional information. The maximum number of available epochs for training is set as $300$ to explore when the model can converge, which costs about $92.5$ hours in our environment. The convergence is observed to emerge after $\simeq 150$ epochs. Therefore, users can consider a $150$-epoch training in the application, which costs about $2$ days under the same hardware condition. There is no additional trick applied during training. 

In \textbf{Fig. 4}, we visualize several instances of the predicted climate fields during testing, accompanied by corresponding true values. High consistency can be seen between these ground truths and their predicted counterparts derived by ViT-KNO. Quantitatively, the prediction accuracy of each climate variable during testing is measured by anomaly correlation coefficient (ACC) in \textbf{Fig. 5}. According to the same prediction task reported by Refs. \cite{pathak2022fourcastnet,kurth2022fourcastnet}, the trained ViT-KNO outperforms the baseline state-of-the-art deep learning models for weather forecasting proposed by Ref. \cite{weyn2021sub} significantly. Compared with the FourCastNet trained with multiple numerical tricks (e.g., multi-stage training with large memory consumption) \cite{pathak2022fourcastnet,kurth2022fourcastnet}, ViT-KNO achieves a similar accuracy during testing. Limited by computing resources, we are unable to precisely compare between ViT-KNO and FourCastNet under the same hardware and training condition yet (the FourCastNet is reported to be trained on 3808 NVIDIA A100 GPUs with numerous computational optimization \cite{pathak2022fourcastnet}). More detailed comparisons may be considered in future studies. We suggest that ViT-KNO has the potential to become a competitive alternative of FourCastNet. Moreover, the time cost of a single time of prediction by ViT-KNO is observed as $\simeq 0.02768695354$ seconds in a single 16-GB GPU. Compared with the classic numerical weather forecasting systems (e.g., the ECMWF Integrated Forecasting System) whose prediction inevitably requires a multi-GPU environment (e.g., more than $1000$ NVIDIA Selene nodes where each node consists of $8$ NVIDIA A100 GPUs) \cite{roberts2018climate,lopez2016lightning,bauer2020ecmwf}, ViT-KNO is orders of magnitude faster in the application (e.g., the Integrated Forecasting System L91 18
km model is expected to cost about $9840$ node seconds for prediction on a NVIDIA Selene node \cite{bauer2020ecmwf}). Therefore, our ViT-KNO has the potential to serve as a unit in ensemble weather forecasting frameworks to realize an efficient prediction of global weather. 

\section{Conclusion}
In this paper, we have presented KoopmanLab, an efficient module of Koopman neural operator family for solving partial differential equations. The included models in this module, such as the compact KNO sub-family and the ViT-KNO sub-family, are provided with mathematical foundations, computational designs, and validations in solving concrete PDEs or predicting intricate dynamic system governed by unknown coupled PDEs. All models are suggested as competitive with other state-of-the-art approaches in corresponding tasks. Compared with classic numerical and neural-network-based PDE solvers, the proposed KNO variants can achieve significant acceleration, more robust mesh-independence, higher generalization capacity on changed conditions, more flexibility in characterizing latent PDEs with unknown forms, and a better balance between accuracy and efficiency. Therefore, we suggest the potential of KoopmanLab be applied in diverse down-stream tasks related with PDE solving. Users can download this module via
 \begin{lstlisting}[language=Python]
pip install koopmanlab
\end{lstlisting}
or 
 \begin{lstlisting}[language=Python]
git clone https://github.com/Koopman-Laboratory/KoopmanLab.git
cd KoopmanLab
pip install -e .
\end{lstlisting}

Several important questions remain for future studies. First, one may consider more specialized computational optimization of models in KoopmanLab (e.g., consider multi-stage training as suggested in Refs. \cite{pathak2022fourcastnet,kurth2022fourcastnet} or multi-objective balancing by Pareto theory \cite{sener2018multi,momma2022multi}). Second, one can explore a more detailed comparison between the ViT-KNO sub-family and FourCastNet under the equivalent hardware and training conditions. Third, one can analyze the errors of our models caused by the potential continuous spectrum of the Koopman operator or the absence of ergodic property in real cases.

\section*{Acknowledgements}
This project is supported by the Artificial and General Intelligence Research Program of Guo Qiang Research Institute at Tsinghua University (2020GQG1017) as well as the Tsinghua University Initiative Scientific Research Program. 

\bibliography{apssamp}
\end{document}